%% file: main.tex
\documentclass[11pt]{article}

\usepackage[a4paper, left=1in, right=1in, top=1.2in, bottom=1.2in]{geometry}
 
\usepackage{setspace}
\usepackage{hyperref}   
\usepackage{float}      
\usepackage{placeins}   

\usepackage{graphicx}       
\usepackage{subcaption}     
\usepackage{wrapfig}        
\usepackage{booktabs}       
\usepackage{multirow}       
\usepackage{caption}        
\usepackage{pgfplots}       
\pgfplotsset{compat=1.18}
\usepackage{tikz}           

\usepackage{amsmath, amssymb, amsthm, dsfont} 
\usepackage{dsfont}
\usepackage{algorithm}
\usepackage{algpseudocode}

\usepackage{enumitem}       

\newtheorem{theorem}{Theorem}[section]  

\theoremstyle{definition}

\title{Fair Supervised Learning Through Constraints on Smooth Nonconvex Unfairness-Measure Surrogates}

\author{
Zahra Khatti\thanks{Department of Industrial and Systems Engineering, Lehigh University. Email: zak223@lehigh.edu} 
\and
Daniel P. Robinson\thanks{Department of Industrial and Systems Engineering, Lehigh University. Email: daniel.p.robinson@lehigh.edu}
\and
Frank E. Curtis\thanks{Department of Industrial and Systems Engineering, Lehigh University. Email: frank.e.curtis@lehigh.edu}
}
\date{}
\begin{document}
\maketitle

\begin{abstract}
    
   A new strategy for fair supervised machine learning (ML) is proposed.  Its advantages compared to others are as follows. (a) We introduce a new smooth nonconvex surrogate to approximate the Heaviside functions involved in discontinuous unfairness measures.  The surrogate is based on smoothing methods from the optimization literature, and is new for fair supervised ML. The surrogate is a tight approximation that ensures the trained prediction models are fair, as opposed to other (e.g., convex) surrogates that can fail to lead to fair prediction models. (b) Rather than rely on regularizers (that lead to optimization problems that are difficult to solve) and corresponding regularization parameters (that can be expensive to tune), we propose a strategy that employs hard constraints so that specific tolerances for unfairness can be enforced without the complications associated with the use of regularization.  (c)~Our strategy readily allows for constraints on multiple (potentially conflicting) unfairness measures at the same time.  Multiple measures can be considered with a regularization approach, but at the cost of having even more difficult training problems and further expense for tuning.  By contrast, through hard constraints, our strategy leads to training problems that can be solved tractably through minimal tuning.
   \end{abstract}

\section{Introduction}

Prediction models that have been created using supervised machine learning (ML) techniques are very effective in modern practice.  For a few examples relevant to this paper, we mention that researchers and practitioners have witnessed impressive performance by such models in finance~\cite{fuster2022predictably}, criminal justice~\cite{berk2021fairness}, hiring~\cite{raghavan2020mitigating, dastin2022amazon}, and healthcare~\cite{Obermeyer2019health}.  However, it has also been found that such prediction models can be affected by and/or introduce biases that can lead to unfair decisions.  This can have discriminatory and legal implications \cite{barocas2016big}, such as when certain demographic groups receive favorable or unfavorable outcomes \cite{feldman2015certifying} to a disproportionate degree compared to other groups.  Overall, these biases---either inherited from historical data \cite{holstein2019improving, lum2016predict} or resulting from model training procedures \cite{zhao2017men, holstein2019improving}---raise significant concerns about the use of supervised ML models in real-world applications.  For only a few examples: Research has shown that Black individuals are often offered higher interest rates for auto loans \cite{charles2008rates}, small business loans for women can involve higher interest rates than for men \cite{alesina2013women}, and gender biases persist in algorithmic hiring tools \cite{dastin2022amazon}.

Various strategies have been explored that aim to produce prediction models that may be considered fair.  Broadly characterized, these are referred to as pre-processing strategies that involve modifying training data before model training (see, e.g., \cite{kamiran2012data}), in-processing strategies that involve modifying training algorithms in order to produce fair models (see, e.g., \cite{zafar2017fairconst, agarwal2018reductions}), and post-processing strategies that adjust a model after it has been trained (see, e.g., \cite{pleiss2017fairness}).  Our focus in this work is to devise a new in-process strategy.  In particular, we focus our attention on the incorporation of unfairness measures in the optimization problems that arise for model training, since such an approach has been shown to lead to promising improvements \cite{friedler2019comparative}.  (The literature refers to both fairness measures and unfairness measures.  We refer to unfairness measures in this paper since they lead to quantities that we aim to constrain, i.e., to have low levels of unfairness.)

The use of unfairness measures in an optimization problem that is formulated for model training has been explored previously.  However, due to computational challenges that arise from the incorporation of such measures, compromises are often made that can diminish the effectiveness of such an approach.  For one thing, the straightforward formulation of an explicit unfairness measure can lead to a function that is nonconvex and/or nonsmooth, which in turn causes computational challenges in the empirical risk minimization (ERM) problems that are formulated for model training \cite{vapnik1999nature}.  As a result, convex surrogates \cite{zafar2017fairconst, donini2018empirical}---such as those based on covariance approximations \cite{goh2016satisfying}---are often employed with the aim of making the ERM problem easier to solve.  However, a downside of this approach is that while it may ensure that a covariance measure (between predictions and a sensitive feature) is reduced, it does not guarantee that the original unfairness measure is kept within reasonable limits.  Second, the widespread availability of software that typically focuses on (stochastic-)gradient-based algorithms often leads to the use of unfairness measures only through regularization terms in the objective function \cite{kamishima2012fairness, bechavod2017penalizing}.  However, this leads to formulations that are expensive and time-consuming to tune, and may ultimately fail to enforce fairness strictly \cite{celis2019classification}.  Thus, the model may possess residual biases and lead to disparate outcomes \cite{woodworth2017learning}.

Previously proposed strategies face other challenges as well.  For instance, some strategies face difficulties when trying to incorporate multiple unfairness measures that conflict with each other \cite{kleinberg2017inherent, chouldechova2017fair}.  This adds to the fundamental challenge of balancing accuracy and fairness \cite{menon2018cost}, where---if one is not careful---model performance can suffer when trying to enforce a limit on an unfairness measure \cite{friedler2019comparative}.  In addition, some strategies---say, involving relaxed constraints \cite{goh2016satisfying} or adversarial techniques \cite{zhang2018mitigating}---can struggle when trying to balance computational efficiency with fairness guarantees and model performance \cite{martinez2020minimax}.

\subsection{Contributions}

We propose a new in-process strategy for fair supervised ML with the following contributions.

\begin{itemize}
  \item We propose the use of smooth \emph{nonconvex} and \emph{bounded} empirical models of discontinuous unfairness measures that, through a scalar parameterization, can approximate unfairness measures to arbitrarily high accuracy. Moreover, we prove that the unfairness measure is small whenever our proposed model value is small (e.g., if used as a constraint in an optimization formulation).  This is in sharp contrast to commonly used \emph{convex} or \emph{unbounded} nonconvex surrogates \cite{lohaus2020too, mary2019fairness, wu2018fairness, zafar2017fairconst, zhang2018mitigating}, which do not provide such a guarantee, meaning that the unfairness measure can be large  even when such surrogates are small.  These claims are supported by our numerical experiments with multiple datasets and multiple (convex and nonconvex) surrogate functions.
  \item Rather than incorporate unfairness measures only through regularization terms (otherwise known as soft constraints) \cite{kamishima2012fairness, bechavod2017penalizing, padala2020fnnc}, we propose the use of hard constraints on surrogates of unfairness measures \cite{zafar2017fairconst, donini2018empirical, cotter2019optimization}.  This is the first paper to combine hard constraints with smooth nonconvex surrogates in such a way that desired bounds on unfairness measures are actually realized (to high accuracy) when a prediction model is ultimately trained.  Our approach of using hard constraints significantly reduces training cost, since otherwise a significant amount of computational expense needs to be devoted to tuning regularization parameters.  We also show through our experiments that setting a higher regularization in order to ensure further decrease in an unfairness measure can have a significantly adverse affect on prediction accuracy, whereas by formulating and solving a problem with hard constraints the effect on prediction accuracy can be much more mild.
  \item Formulations that involve only a single unfairness measure might fail to account for different issues that may make a prediction model unfair \cite{hardt2016equality, zafar2017fairconst, zhang2018mitigating}.  In contrast, by employing hard constraints on accurate unfairness-measure surrogates, our approach readily allows the incorporation of multiple (potentially conflicting) unfairness measures at the same time \cite{barocas2017fairness, mary2019fairness, celis2019classification}. This flexibility enhances the capability of our approach to mitigate multiple types of biases, ensuring broader applicability and robustness~\cite{friedler2019comparative} for training fair prediction models.
\end{itemize}

\section{Background on Unfairness Measures}\label{sec.background}  

The purpose of this section is to provide background on unfairness measures that are relevant for our setting.  We begin by referencing a few fundamental fairness criteria that are well known in the supervised learning literature, then present probabilistic statements of fairness concepts that will, in turn, be used to formulate measures of unfairness that we will employ in our subsequent problem formulations and numerical experiments.  We conclude this section with some comments about the use of convex and nonconvex surrogate functions for approximating unfairness measures.

\subsection{Fundamental Fairness Criteria}


Consider the setting of supervised ML for classification, where in contrast to the standard setting of having only a combined feature vector, we distinguish between nonsensitive and sensitive features.  (One idea for producing a fair classifier is to remove the sensitive feature.  However, this runs the risk of producing an unfair classifier since nonsensitive features can be correlated with sensitive ones \cite{grgic2016cas}.)  For the sake of simplicity, consider the case where there is a single sensitive feature that takes binary values and a single label that takes binary values.  Our proposed approach can readily be extended to cases involving any finite number of sensitive feature values and multiclass settings with more than two labels. This can be done through constraints on each pair of sensitive feature and label.

Let $(X, S, Y)$ be a tuple of random variables that is defined with respect to a probability space $(\Omega, \cal{F}, \mathbb{P})$, where $X$ represents the nonsensitive feature vector, $S$ represents the binary sensitive feature (in $\{0,1\}$), and $Y$ represents the binary true label (in $\{0,1\}$).  Through a supervised learning procedure, suppose one defines a prediction function parameterized by $w$ so that, with a \emph{trained} $w$, the predicted label corresponding to a given pair of features is given by $\hat{Y} \equiv \hat{Y}(X, S, w)$.  Three fundamental fairness criteria that are recognized in the literature are independence, separation, and sufficiency \cite{barocas2017fairness}.  The independence criterion requires that the prediction $\hat{Y}$ is independent of the sensitive feature $S$; the separation criterion requires that the prediction $\hat{Y}$ is independent of the sensitive feature $S$ conditioned on the true label $Y$; and the sufficiency criterion requires that the true label $Y$ is independent of the sensitive feature $S$ conditioned on the prediction $\hat{Y}$.

We focus primarily on unfairness measures based on the independence criterion, although our proposed strategy can be extended readily to the context of alternative criteria.

\subsection{Probabilistic Statements of Fairness}\label{sec.MathematicalMeasures}  

From the aforementioned fundamental criteria, various specific probabilistic statements of fairness have been derived \cite{friedler2019comparative}.  Here, we provide two related examples based on the independence criterion.

\paragraph{Demographic Parity (or Statistical Parity).}

Demographic parity requires that a model's predictions are independent of the sensitive attribute \cite{zemel2013learning}:
{\small
\begin{equation*}
  \mathbb{P}(\hat{Y}(X, S, w) = 1 | S = 1) = \mathbb{P}(\hat{Y}(X, S, w) = 1 | S = 0).
\end{equation*}}
Later on, we will discuss the empirical violation of this equation as an unfairness measure.

\paragraph{Disparate Impact.}

Disparate impact relates to the requirement that selection rates are similar regardless of the value of the sensitive feature.  The aim to avoid disparate impact can be seen, e.g., in the ``four-fifths rule'' in US employment law \cite{barocas2016big,feldman2015certifying,fourfifths}.  Stated mathematically:
{\small
\begin{equation*}
  \begin{aligned}
    \delta \mathbb{P}(\hat{Y}(X, S, w) = 1 | S = 1) &\leq \mathbb{P}(\hat{Y}(X, S, w) = 1 | S = 0) \\
    \delta \mathbb{P}(\hat{Y}(X, S, w) = 1 | S = 0) &\leq \mathbb{P}(\hat{Y}(X, S, w) = 1 | S = 1),
  \end{aligned}
\end{equation*}}
where $\delta \in [0,1]$ is a threshold parameter (e.g., $\delta = 0.8$ in the context of the four-fifths rule).  Later on, we will employ empirical approximations of the terms in these inequalities in order to formulate constraints (on smooth nonconvex surrogates) to be imposed during model training.

\paragraph{Equal Impact.} This requires that predictions given a positive label are independent of the sensitive attribute:
{\small
\begin{equation*}
  \begin{aligned}
    \delta \mathbb{P}(\hat{Y}(X, S, w) = 1 | Y = 1, S = 1) 
    \leq \mathbb{P}(\hat{Y}(X, S, w) = 1 | Y = 1, S = 0) \\
    \delta \mathbb{P}(\hat{Y}(X, S, w) = 1 | Y = 1, S = 0) 
    \leq \mathbb{P}(\hat{Y}(X, S, w) = 1 | Y =1, S = 1)
  \end{aligned}
\end{equation*}}
where $\delta \in [0,1]$ is a threshold parameter.

\subsection{Empirical Surrogates for Probabilistic Statements}

These probabilistic statements of fairness can be incorporated into optimization problem formulations designed for prediction model training if and only if they are replaced by empirical approximations. Suppose that $w$ represents trainable parameters, such as of a deep neural network, and that a set of feature-label tuples $\{(x_i,s_i,y_i)\}_{i\in[N]}$ is available for prediction model training, where $[N] := \{1,\dots,N\}$.  For each $i \in [N]$, let the output of the neural network be given by ${\cal N}(x_i,s_i,w)$ and suppose that, for a given threshold $\tau \in \mathbb{R}$ and with $\mathds{1}$ denoting the indicator function, the predicted label for $(x_i,s_i)$ is given by
\begin{equation*}
  \hat{y}(x_i, s_i, w) = \mathbf{1} \{ {\cal N}(x_i, s_i, w) > \tau \}.
\end{equation*}
For each value $s \in \{0,1\}$, the probability of a positive prediction over group $s$ can be estimated as
\begin{equation*}
  \mathbb{P}(\hat{Y}(X, S, w) = 1 | S = s) \approx \frac{\sum_{\{i \in [N] : s_i = s\}} \hat{y}(x_i, s_i, w)}{N_s},
\end{equation*}
where $N_s := \sum_{i \in [N]} \mathbf{1} \{s_i = s\}$.  However, since the indicator function is discontinuous, it cannot be used directly to formulate an optimization problem that is to be solved using an algorithm for continuous optimization \cite{woodworth2017learning}.  Thus, as is common in the literature, let us suppose that for all $i \in [N]$ one approximates
\begin{equation}\label{def:t}
  \begin{aligned}
    \hat{y}(x_i, s_i, w) &\approx \phi(t(x_i,s_i,w)), \text{where}\ \ t(x_i,s_i,w) := {\cal N}(x_i,s_i,w) - \tau.
  \end{aligned}
\end{equation}
Here, $t(x_i,s_i,w)$ represents the distance-to-threshold function for data point $i \in [N]$ and $\phi : \mathbb{R} \to \mathbb{R}$ approximates the step function from 0 to 1 (at the origin).  Various such approximations have been explored for approximating fairness measures and other purposes as well.  To list a few, we mention the linear function $\phi(t) = t$ \cite{zafar2017fairconst}, the max function $\phi(t) = \max\{t,0\}$ \cite{wu2019convexity}, the sigmoid function $\phi(t) = \sigma(t) := (1 + e^{-t})^{-1}$ \cite{bendekgey2021scalable}, and the hyperbolic tangent function $\phi(t) = \tanh(t)$.  We focus on the sigmoid function and another approximation---based on a smoothing function that is common in the optimization literature---for use in the formulations that we propose for prediction model training.

A question that arises in the use of a surrogate to approximate a step function is how large discrepancies can be once the model is trained.  The following theorem, derived from~\cite{yao2023understanding}, establishes a revealing bound on such discrepancies \emph{when the codomain of $\phi$ is the unit interval and the shifted function $\phi-\tfrac12$ is symmetric about the origin}; a proof is given in Appendix A to account for minor changes in the properties of $\phi$ as compared to~\cite{yao2023understanding}. It is important to note that while the corresponding theorem from \cite{yao2023understanding} refers to bounds on absolute values of differences of empirical estimates of probabilities, the numerical studies in that paper involve regularization for model training, meaning that such constraints were not necessarily satisfied in practice.  This further motivates a unique feature of our work, which shows that by enforcing hard constraints we can meet the useful bounds stated in this theorem.

\begin{theorem}\label{th.surrogate}
  (derived from \cite{yao2023understanding}) Suppose $\phi : \mathbb{R} \to [0,1]$, the function $\phi - \tfrac12$ is symmetric about the origin, and for some $\gamma \in (0,\tfrac12)$ one has $\phi(t(x_i,s_i,w)) \in [0,\gamma] \cup [1-\gamma,1]$ for all $i \in [N]$.  Suppose also that, for some $\epsilon \in (0,\infty)$, the surrogate of the empirical estimate of the violation of demographic parity has $|c_{\textrm{dp}}(w)| \leq \epsilon$ where
  \begin{equation}\label{eq.dp}
    \begin{aligned}
      c_{\textrm{dp}}(w) :=\ \frac{\sum_{\{i \in [N] : s_i = 1\}} \phi(t(x_i, s_i, w))}{N_1} \ - \frac{\sum_{\{i \in [N] : s_i = 0\}} \phi(t(x_i, s_i, w))}{N_0}.
    \end{aligned}
  \end{equation}
  Then the actual empirical estimate of the violation of demographic parity has $|\bar{c}_{\textrm{dp}}(w)| \leq \epsilon + \gamma$ where
  \begin{equation*}
    \begin{aligned}
      \bar{c}_{\textrm{dp}}(w) :=\ \frac{\sum_{\{i \in [N] : s_i = 1\}} \hat{y}(x_i,s_i,w)}{N_1} \ - \frac{\sum_{\{i \in [N] : s_i = 0\}} \hat{y}(x_i,s_i,w)}{N_0}.
    \end{aligned}
  \end{equation*}
\end{theorem}

Let us close this section by mentioning that the common covariance-based surrogate for independence \cite{zafar2017fairconst} can be understood in terms of a specific surrogate of the form mentioned above.  A benefit of this surrogate is that it is convex.  However, a downside is that it does not readily offer a means to leverage the useful result in Theorem~\ref{th.surrogate}.  Specifically, the surrogate is given by

{\small
\begin{align*}
  &\textrm{cov}(\hat{Y}(X,S,w), S)
   = \mathbb{E}[(\hat{Y}(X,S,w) - \mathbb{E}[\hat{Y}(X,S,w)])(S - \mathbb{E}[S])] \\
  &\approx \frac{1}{N} \sum_{i \in [N]} (s_i - \bar{s}) \cdot {\cal N}(x_i, s_i, w) 
   = c_{\textrm{cov}}(w).
\end{align*}}

where $\bar{s} = \frac{1}{N} \sum_{i \in [N]} s_i$.  The following theorem \cite{bendekgey2021scalable,yao2023understanding} shows that this surrogate is proportional to $c_{\textrm{dp}}$ from \eqref{eq.dp} with the specific choice of $\phi(t) = t$ (linear function), namely,
{\small
\begin{align*}
  &\mathbb{E}[\hat{Y}(X,S,w) \mid S = 1] - \mathbb{E}[\hat{Y}(X,S,w) \mid S = 0] \\
  &\approx \frac{\sum_{\{i \in [N] : s_i = 1\}} t(x_i, s_i, w)}{N_1} - \frac{\sum_{\{i \in [N] : s_i = 0\}} t(x_i, s_i, w)}{N_0} 
   = c_{\textrm{dp}}^{\phi(t)=t}(w).
\end{align*}}

\begin{theorem}\label{th.cov}
  The functions $c_{\textrm{cov}}$ and $c_{\textrm{dp}}^{\phi(t)=t}$ satisfy $c_{\textrm{cov}}(w) = \tfrac{N_0 \cdot N_1}{N^2} \cdot c_{\textrm{dp}}^{\phi(t)=t}(w)$ for all $w$.
\end{theorem}

The proportionality shown in Theorem~\ref{th.cov} suggests that one might be able to enforce a bound on the violation of demographic parity by enforcing $|c_{\textrm{cov}}(w)| \leq \epsilon$ for some $\epsilon \in (0,\infty)$.  However, the useful bound in Theorem~\ref{th.surrogate} is not readily applicable to this setting since the codomain of the linear function $\phi$ defined by $\phi(t) = t$ is unbounded; thus, it is essentially impossible to ensure that the surrogate yields, for some $\gamma \in (0,\tfrac12)$ and all $i \in [N]$, the inclusion $\phi(t(x_i,s_i,w)) \in [0,\gamma] \cup [1-\gamma,1]$ for all points in the training set.  Therefore, $|c_{\textrm{cov}}(w)| \leq \epsilon$ does not necessarily enforce a useful bound on the actual empirical estimate of violation of demographic parity, which motivates our choice of studying more accurate, nonconvex and bounded surrogates.

\section{Methodology}\label{sec.methodology}

Motivated by our prior discussions, we now present our proposed methodology that combines smooth, nonconvex, and accurate surrogate functions for unfairness measures with hard constraints that are to be imposed during model training.  We emphasize that this combination leads to a \emph{tractable} training paradigm where the optimization problem can be solved using stochastic Newton/SQP techniques from the recent literature \cite{curtis2023fair}.  By \emph{tractable}, we are emphasizing the fact that one can solve a constrained training problem to obtain a prescribed limit on unfairness by solving a single optimization problem.  Naturally, this optimization problem is more expensive to solve than a single unconstrained problem that naively introduces an unfairness measure through an objective regularization term.  However, overall, there are computational savings due to the fact that, by solving a constrained problem, one does not need to tune regularization parameters for the unfairness measure terms.  We also emphasize that \cite{curtis2023fair} shows that constrained training problems can be solved efficiently despite nonconvexity in both the objective and constraint functions.  The benefits of our methodology are realized in practice in the experimental results that we present in the following section.

\subsection{Smooth, Nonconvex, and Accurate Surrogates}\label{sec.scaling}

At the heart of the definition of a surrogate function for our setting is an approximation of the step function from 0 to~1 (at the origin), also known as the Heaviside function.  In particular, due to Theorem~\ref{th.surrogate}, we are interested in bounded approximations, say, ones with a codomain of $[0,1]$.  A common choice with these properties is the sigmoid function, so that is one of the choices that we consider for our methodology.  However, we also consider a second approximation that employs ideas from smoothing techniques from the mathematical optimization literature; see, e.g., \cite{Chen2012}.  In our experiments, we often find better results for this second approximation---which we refer to as the smoothed-step function---as compared to the sigmoid function.

A piecewise-affine approximation of the step function that, as required by Theorem~\ref{th.surrogate}, has a graph that passes through $(0,0.5)$ is given by $\phi : \mathbb{R} \to [0,1]$ with $\phi(t) = \min\{\max\{0,t + 0.5\},1\}$ for all $t \in \mathbb{R}$.  This can be viewed as a composite involving two max's: $\phi(t) = \underline\phi(\overline\phi(t))$ where
\begin{equation*}
  \begin{aligned}
  \underline\phi(\bar{t}) = \min\{\bar{t},1\} = 1 - \max\{1 - \bar{t},0\} \text{and}\ \ \overline\phi(t) = \max\{0,t + 0.5\}.
  \end{aligned}
\end{equation*}
Now employing the smooth approximation given by $\max\{t,0\} \approx \tfrac12 (t + \sqrt{t^2 + \mu})$, where $\mu \in (0,\infty)$ is user-defined, one obtains the smooth, nonconvex approximation
\begin{align*}
  & \phi_\mu(t(x_i,s_i,w)) :=
  \ 1 - \tfrac{1}{2} \bigg( 1 - \overline\phi_\mu(t(x_i,s_i,w)) \ + \sqrt{(1 - \overline\phi_\mu(t(x_i,s_i,w)))^2 + \mu} \bigg)\\
  &  \text{where}\ \ \overline\phi_\mu(t(x_i,s_i,w))
  :=\ \tfrac{1}{2} \left( t(x_i, s_i, w) + \tfrac12 + \sqrt{(t(x_i, s_i, w) + \tfrac12)^2 + \mu} \right).
\end{align*}

\begin{figure}[ht]
  \centering
  \begin{subfigure}[b]{0.49\textwidth} 
    \centering
    \includegraphics[width=\textwidth]{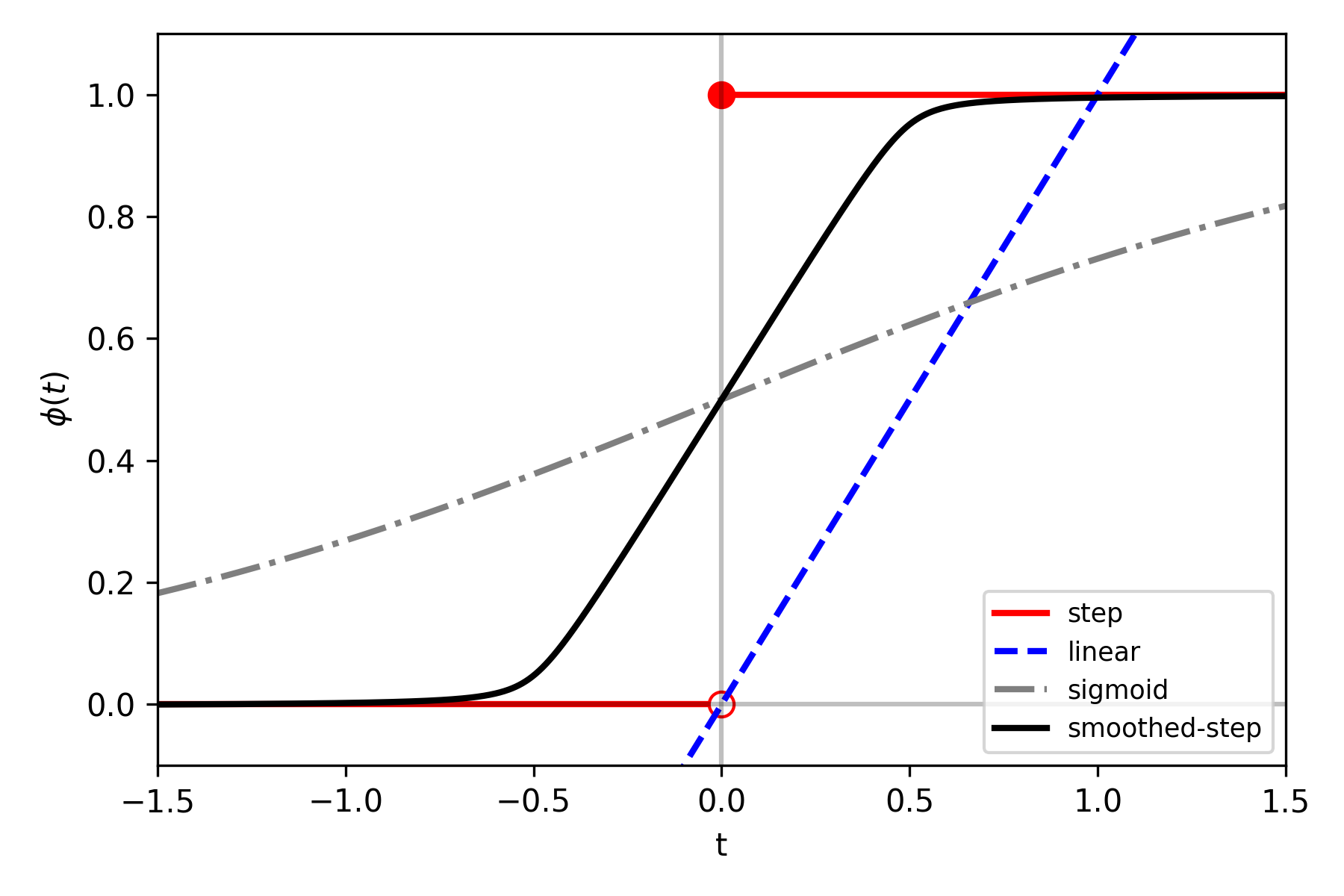}
    \caption{Surrogate functions}
  \end{subfigure}
\hfill
  \begin{subfigure}[b]{0.49\textwidth}
    \centering
    \includegraphics[width=\textwidth]{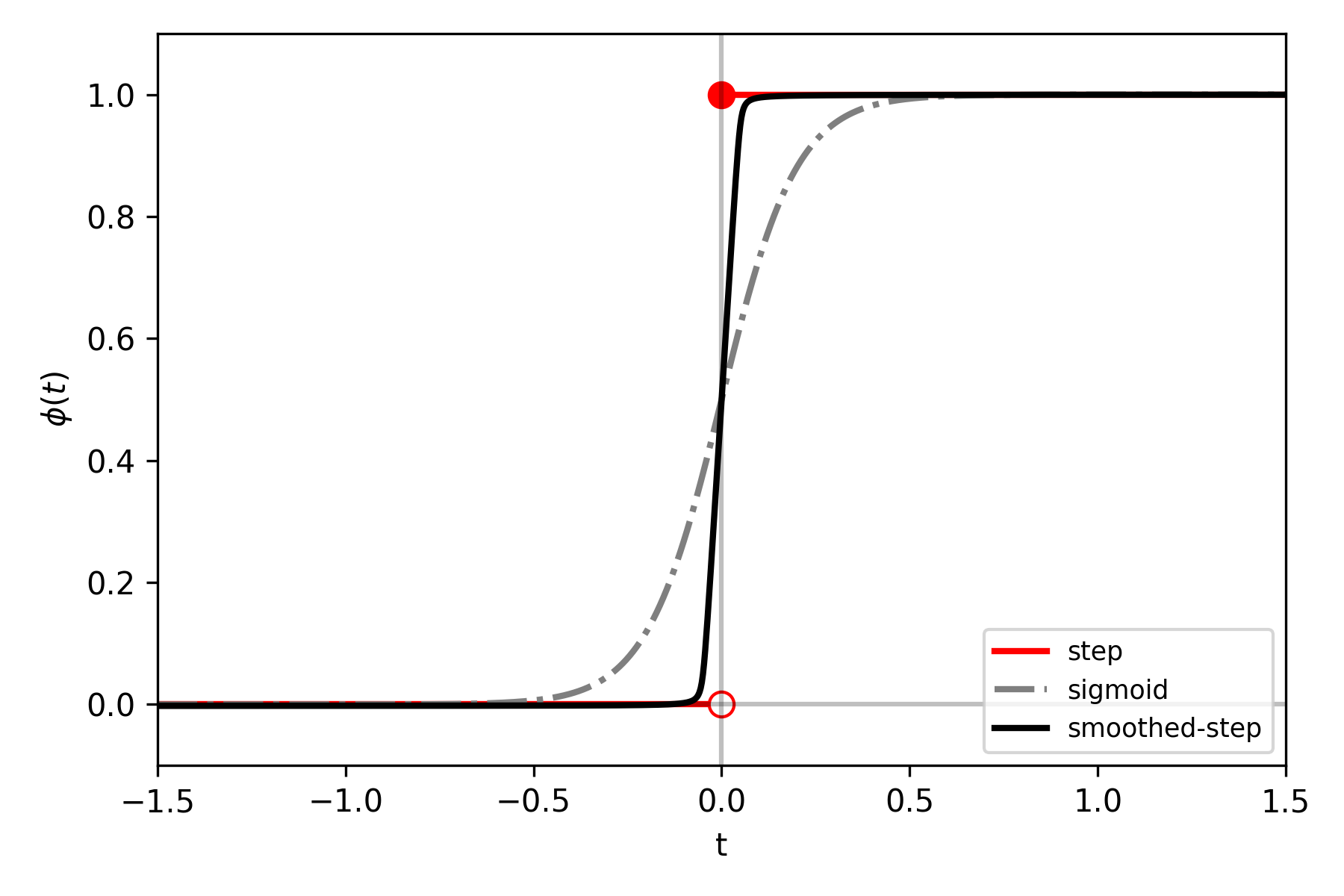}
    \caption{Scaled surrogate functions}
  \end{subfigure}
  \caption{On the left, graphs of the step/Heaviside function ($\mathbf{1}\{t\geq 0\}$), linear function ($\phi(t) = t$), sigmoid function ($\phi(t) = \sigma(t)$), and smoothed-step function ($\phi_\mu(t)$ defined in this section).  On the right, graphs of scaled functions ($\sigma(10t)$ and $\phi_\mu(10t)$) to illustrate that scaling can make the functions more closely approximate the step function.}
  \label{fig:activation-functions}
\end{figure}

Figure~\ref{fig:activation-functions} shows the step function (Heaviside function) along with linear, sigmoid, and smoothed-step approximations.  We also want to emphasize that, to obtain improved results in practice, these functions should be scaled in order to encourage values that are closer to 0 or 1~\cite{yao2023understanding}; in particular, in place of $\sigma(t)$ and $\phi_\mu(t)$ one should consider $\sigma(\alpha t)$ and $\phi_\mu(\alpha t)$ for $\alpha \in (0,\infty)$, likely greater than~1.  Such scalings make the derivative of the function larger near the origin.  Overall, such scaling is beneficial in terms of Theorem~\ref{th.surrogate}, but one might wonder if it will make the training problem more difficult to solve.  We did not find this to be the case with our proposed algorithm, discussed next.  Indeed, overall, our experiments demonstrate improved results with larger scaling values.

\subsection{Training Problem with Unfairness Constraints}\label{sec.fairness_constraints}

Given a set of feature-label tuples $\{(x_i,s_i,y_i)\}_{i\in[N]}$, prediction function~${\cal N}$, loss function~$\ell$, regularization function~$r$, parameter $\lambda \in (0,\infty)$, constraint function $c$, and bounds $(l,u)$, we propose that a fair prediction model can be trained in a supervised ML context by solving the following \emph{tractable} optimization problem with unfairness constraints:
\begin{equation}\label{opt-prob}
  \begin{aligned}
    \min_w\ &\ \sum_{i \in [N]} \ell({\cal N}(x_i,s_i,w),y_i) + \tfrac{1}{\lambda} r(w) \\ \text{subject to} &\ l \leq c(w) \leq u.
  \end{aligned}
\end{equation}
For example, ${\cal N}$ may be defined by a deep neural network with $w$ as the trainable parameters, $\ell$ can be any typical loss function for classification, $r$ may be a sparsity-promoting regularizer or one aimed to penalize a prescribed surrogate for an unfairness measure, and $c$ may be a vector-valued function corresponding to a surrogate for an unfairness measure (or surrogates for multiple unfairness measures).  Problem~\eqref{opt-prob} is nonconvex in general due to the structure of multilayer neural networks.  In our experiments in the next section, we consider for illustrative purposes a regularization function based on the desire to impose demographic parity, namely, $r_{\textrm{dp}}(w) = |q(w)|^2$ where
\begin{equation}\label{eq:dp}
  \begin{aligned}
  q(w) = \frac{\sum_{\{i \in [N] : s_i = 1\}} \phi(t(x_i, s_i, w))}{N_1}  - \frac{\sum_{\{i \in [N] : s_i = 0\}} \phi(t(x_i, s_i, w))}{N_0},
  \end{aligned}
\end{equation}
$\phi$ is a prescribed approximation of the step/Heaviside function, and $t$ is defined as in \eqref{def:t}.  Given a trained prediction function determined by $w$, one can evaluate its performance through $\bar{r}_{\textrm{dp}}(w)$, which is evaluated in the same way as $r_{\textrm{dp}}(w)$ except with $\phi(t(\cdot))$ replaced by $\hat{y}(\cdot)$.  As for the constraint function $c$, we focus primarily on one to constrain disparate impact.  Specifically, given $\delta \in [0,1]$, we consider $l = (-\infty,-\infty)$, $u = (0,0)$, and $c = (c_{\textrm{di},1},c_{\textrm{di},2})$ where
{\small
\begin{equation}\label{eq:di}
  \begin{aligned}
    & c_{\textrm{di},1}(w) = \delta \left( \frac{\sum_{\{i \in [N] : s_i = 0\}} \phi(t(x_i, s_i, w))}{N_0} \right) - \frac{\sum_{\{i \in [N] : s_i = 1\}} \phi(t(x_i, s_i, w))}{N_1} \\ 
    & \text{and}\ \ 
    c_{\textrm{di},2}(w) = \delta \left( \frac{\sum_{\{i \in [N] : s_i = 1\}} \phi(t(x_i, s_i, w))}{N_1} \right)  - \frac{\sum_{\{i \in [N] : s_i = 0\}} \phi(t(x_i, s_i, w))}{N_0}.
  \end{aligned}
\end{equation}}
Given a trained prediction function determined by $w$, one can evaluate violation of these constraints through $c_{\textrm{di}}(w) = \max\{c_{\textrm{di},1}(w),c_{\textrm{di},2}(w)\}$, where a violation occurs if and only if this value is greater than 0.  One can also consider $\bar{c}_{\textrm{di}}(w) = \max\{\bar{c}_{\textrm{di},1}(w),\bar{c}_{\textrm{di},2}(w)\}$, where these functions are defined similarly except with $\phi(t(\cdot))$ replaced by $\hat{y}(\cdot)$.

For the sake of tractability, the constraints can be formulated with only a subset of a full training dataset and one can employ stochastic objective gradients.  One needs to be careful with a regularizer defined through~\eqref{eq:dp} in order to ensure that the stochastic gradient estimate is unbiased.  The following informal theorem shows that unbiased stochastic estimates of the gradient of $q$ in \eqref{eq:dp} can be obtained as long as the mini-batches involve the same number of data points with $s_i = 0$ and $s_i = 1$ in all iterations, which is straightforward to enforce; see Appendix B for more details.

\begin{theorem}\label{th.unbiased}
  (Informal) Suppose that $N_0$ points with $s_i = 0$ are chosen uniformly at random and $N_1$ points with $s_i = 1$ are chosen uniformly at random and that these points are used to compute a stochastic estimate of $\nabla q(w_k)$, where $q$ is defined in \eqref{eq:dp}.  Then, the estimate is unbiased.  
\end{theorem}

\section{Experiments}\label{sec.experiments}

Our experiments focus on three benchmark datasets: \emph{Dutch}~\cite{van20012001}, \emph{Law School}~\cite{wightman1998lsac}, and \emph{ACSIncome}~\cite{ding2021retiring}. Detailed statistics for these datasets are provided in Appendix C. Each dataset is split at a ratio of 80:20 for training and testing data. Dutch contains 60,420 data points with 48,336 training points and 12,084 testing points. Each point has 11 features with gender as the sensitive attribute. Law contains 20,798 data points with 16,638 training points and 4,160 testing data points.  Each data point has 11 features with race as the sensitive attribute. The ACSIncome dataset contains 195,665 data points. We use a random subset of 50,000 points, with 40,000 for training and 10,000 for testing. Each data point has 10 features with gender as the sensitive attribute.

Our proposed approach is agnostic to the prediction model.  For our experiments for Dutch, the prediction model was a feed-forward neural network with one hidden layer (with 32 nodes), Leaky ReLU activation at the hidden layers, and sigmoid activation at the output layer.  For Law, the prediction model was a linear neural network with sigmoid activation.  For ACSIncome, the prediction model was a feed-forward neural network with one hidden layer (with 256 nodes), Leaky ReLU activation at the hidden layers, and sigmoid activation at the output layer. In all of the experimental results displayed in this section, the models were trained using a sequential quadratic optimization (SQP) method, Binary Cross-Entropy loss (BCELoss), and 500 epochs with an initial learning rate of 0.5 that was adjusted dynamically during training; see Appendix D for further details about the training algorithm.  The results in this section use full-batch gradients, but Appendix E contains results with stochastic gradients as well, using the result of Theorem~\ref{th.unbiased}.

All results presented here are with respect to the training data.  Appendix E shows that prediction accuracies, unfairness measures, etc.~are similar for all experiments when one considers testing data. Also, for the sake of variety, for each experiment we show results for a different one of the datasets; the remaining results for the other datasets for each experiment are given in Appendix~E.

\subsection{Effect of Surrogate Function on Actual Limit on Unfairness that is Achieved}

\begin{figure}[t]
    \centering
        \begin{subfigure}[b]{\linewidth}
        \centering
        \includegraphics[width=\linewidth]{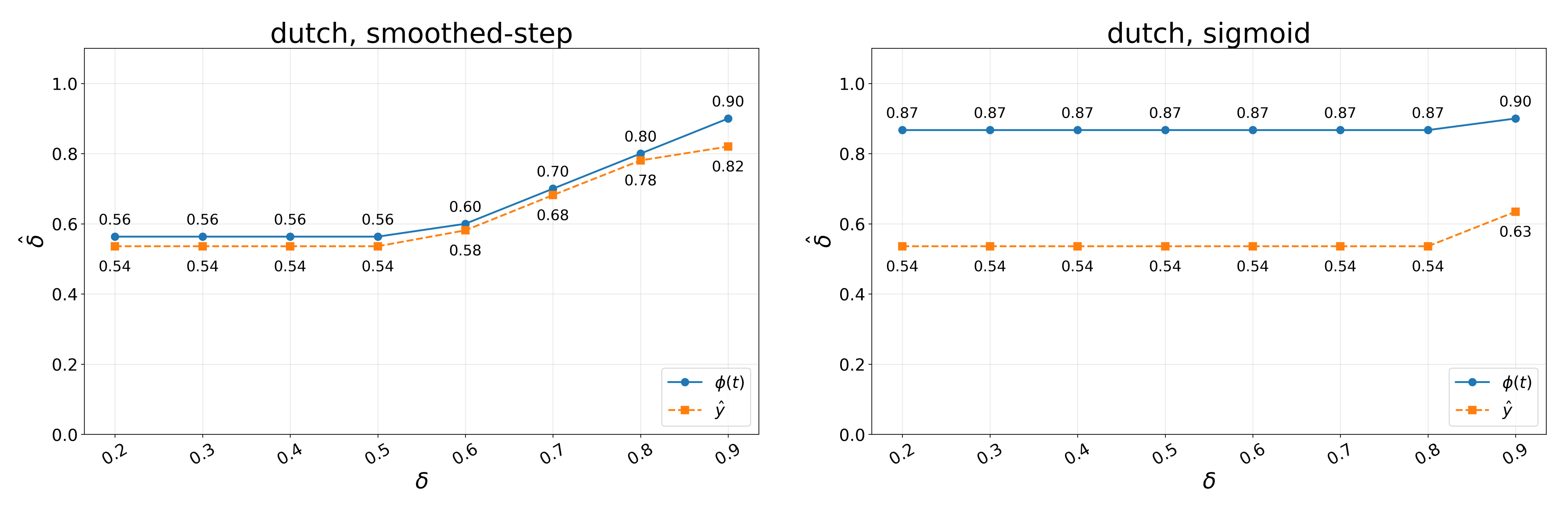}
    \end{subfigure}

    \caption{Levels of disparate impact actually achieved ($\hat\delta$) when prediction models are trained with constraints as in~\eqref{eq:di} for varying values of $\delta$ \emph{without} scaling of the surrogate functions.  The surrogate approximations not being tight causes large gaps between the levels of disparate impact desired and the levels actually achieved.  The graphs indicated by $\phi(t)$ show the values such that $c_{\textrm{di}}(w) \leq 0$, whereas the graphs indicated by $\hat{y}$ show the values such that $\bar{c}_{\textrm{di}}(w) \leq 0$.}
    \label{fig:delta_gap}
\end{figure}

\begin{figure}[t]

    \centering
        
    \begin{subfigure}[b]{\textwidth}
        \centering
        \includegraphics[width=\textwidth]{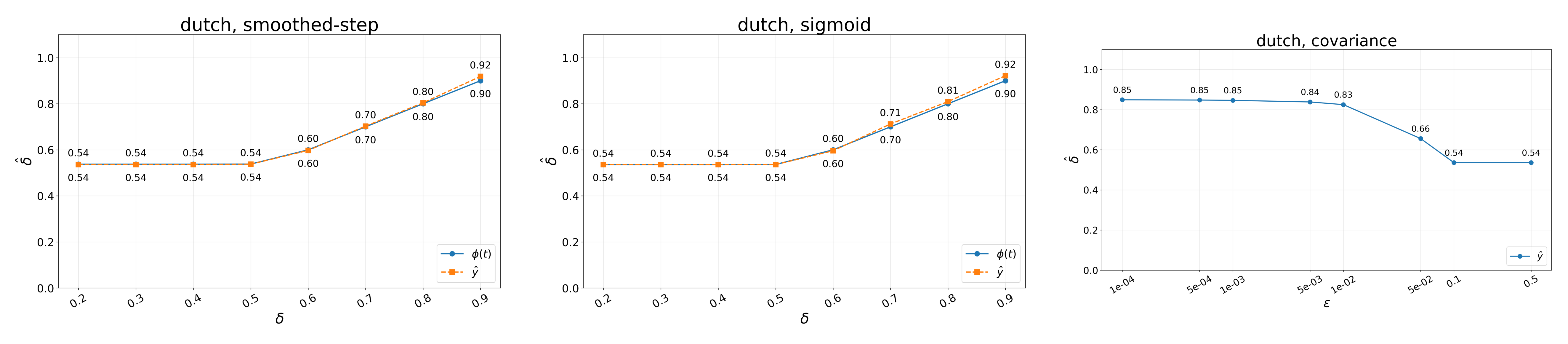}
    \end{subfigure}

    \caption{Levels of disparate impact actually achieved ($\hat\delta$) when prediction models are trained with constraints as in~\eqref{eq:di} for varying values of $\delta$ \emph{with} scaling of the surrogate functions. These results should be contrasted with those in Figure \ref{fig:delta_gap}. In particular, it should be observed that scaling the surrogate functions leads to much tighter correspondence between $\delta$ and $\hat\delta$ when the constraints are tight.  Also, the plot on the right is the levels of disparate impact actually achieved when a constraint on a covariance surrogate (less than or equal to $\epsilon$) is imposed.}
    \label{fig:delta_match}
\end{figure}

Our first set of experiments demonstrates that having a loose approximation of an unfairness measure can cause a trained prediction model to be less fair than desired.  Toward this end, we first trained prediction models using unscaled surrogate functions.  Constraints were imposed as in \eqref{eq:di} for various values of $\delta$.  For each $\delta$, we considered the trained model and determined the maximum value of $\delta$, call it $\hat\delta$, such that $c_{\textrm{di}}(w) \leq 0$ (a similar calculation is also performed based on $\bar{c}_{\textrm{di}}(w) \leq 0$); the results are plotted in Figure~\ref{fig:delta_gap}.  In particular, the graphs labeled $\phi(t)$ and $\hat{y}$ are the plots of the $(\delta, \hat\delta)$ pairs that correspond to $c_{\textrm{di}}(w) \leq 0$ and $\bar{c}_{\textrm{di}}(w) \leq 0$, respectively. These results show that the desired limit on disparate impact was almost achieved when the smoothed-step surrogate was employed, but that the desired limit was not nearly achieved when the sigmoid surrogate was employed.  By contrast, when the surrogate functions are scaled (with $\alpha = 50$) as described previously in this paper, one obtains the results shown in Figure~\ref{fig:delta_match}, where for both the smoothed-step and sigmoid functions one finds that the imposed $\delta$ and the realized $\hat\delta$ values are tightly matched. The scaling parameter ($\alpha = 50$) was selected by running the experiment with $\alpha \in {1, 5, 10, 20, 50, 100}$, then choosing the value that minimized the gap between the levels of fairnes measure desired and the level actually achieved, as shown in Figure~\ref{fig:delta_gap}. Note that these experiments show that the constraints on disparate impact become \emph{active} for Dutch, Law and ACSIncome around $\delta \in [0.5,0.6]$, $\delta \in [0.7,0.8]$, and $\delta \in [0.7,0.8]$, respectively.  In Figure~\ref{fig:delta_match}, we also show that one can impose a constraint on disparate impact through a constraint on a covariance surrogate.  However, in such an approach, it is nontrivial to determine the specific limit on the covariance ($\epsilon$) that corresponds to a specific desired limit on disparate impact.  \emph{This demonstrates that our use of accurate, nonconvex surrogates can make it easier to impose a specific desired limit on an unfairness measure such as disparate impact}.

\begin{figure}[!htbp]
    \centering
    \includegraphics[width=0.48\linewidth]{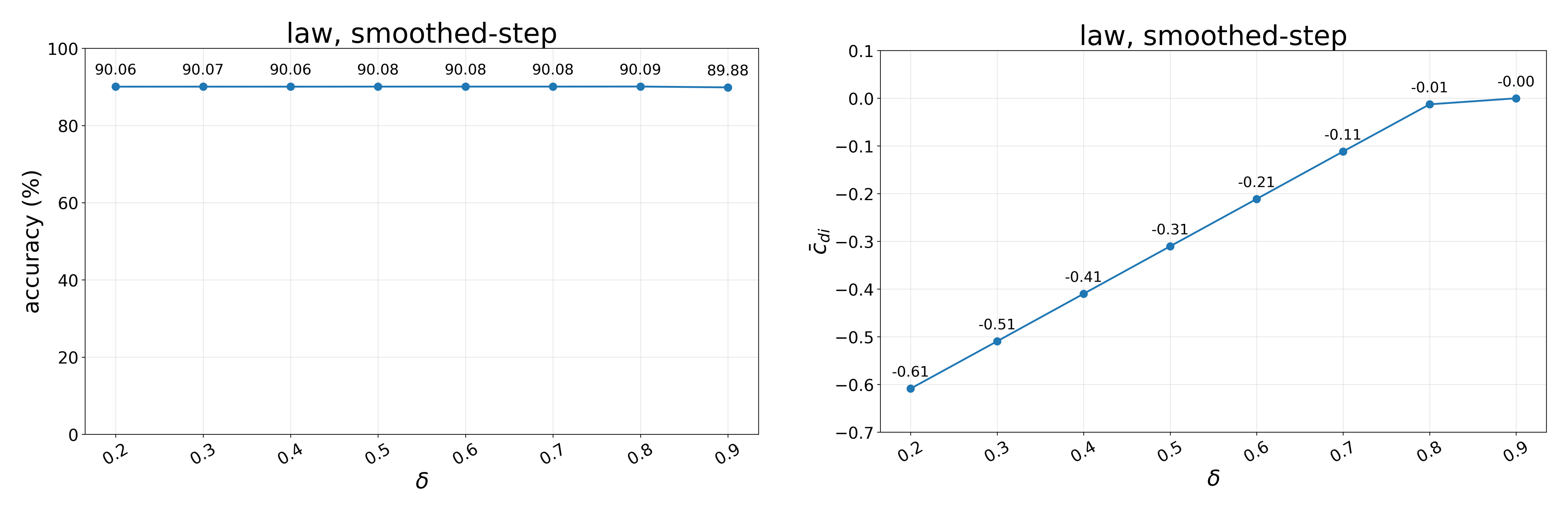}\hfill
    \includegraphics[width=0.48\linewidth]{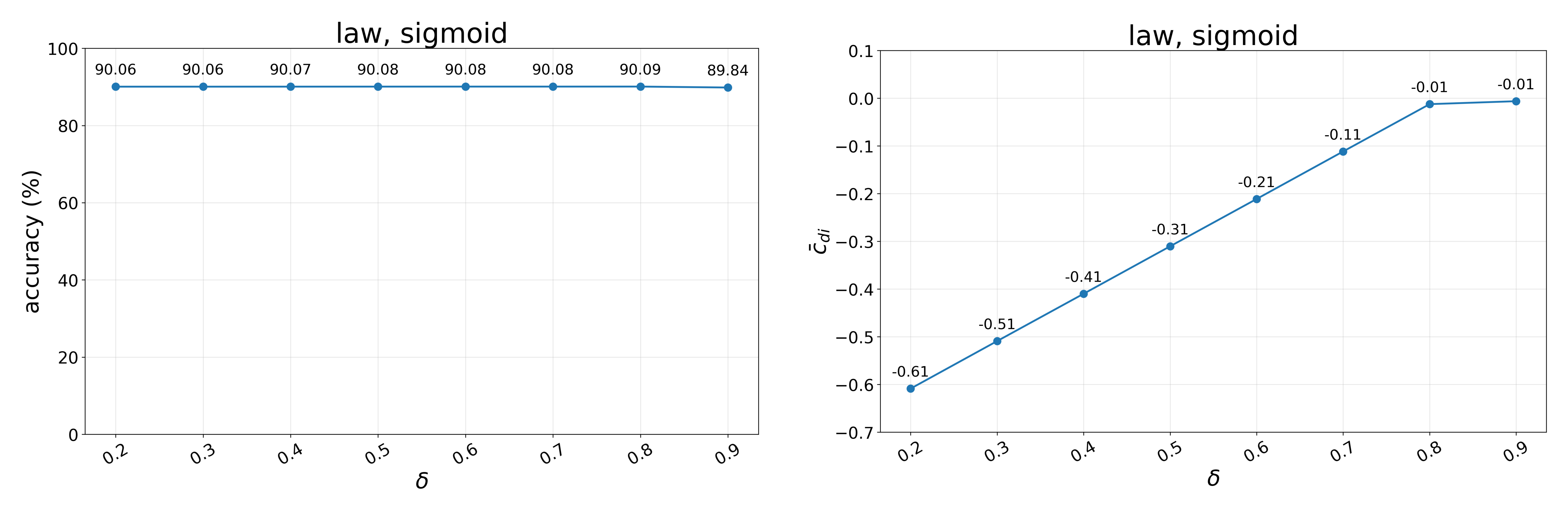}

    \vspace{0.3cm} 

    \includegraphics[width=0.48\linewidth]{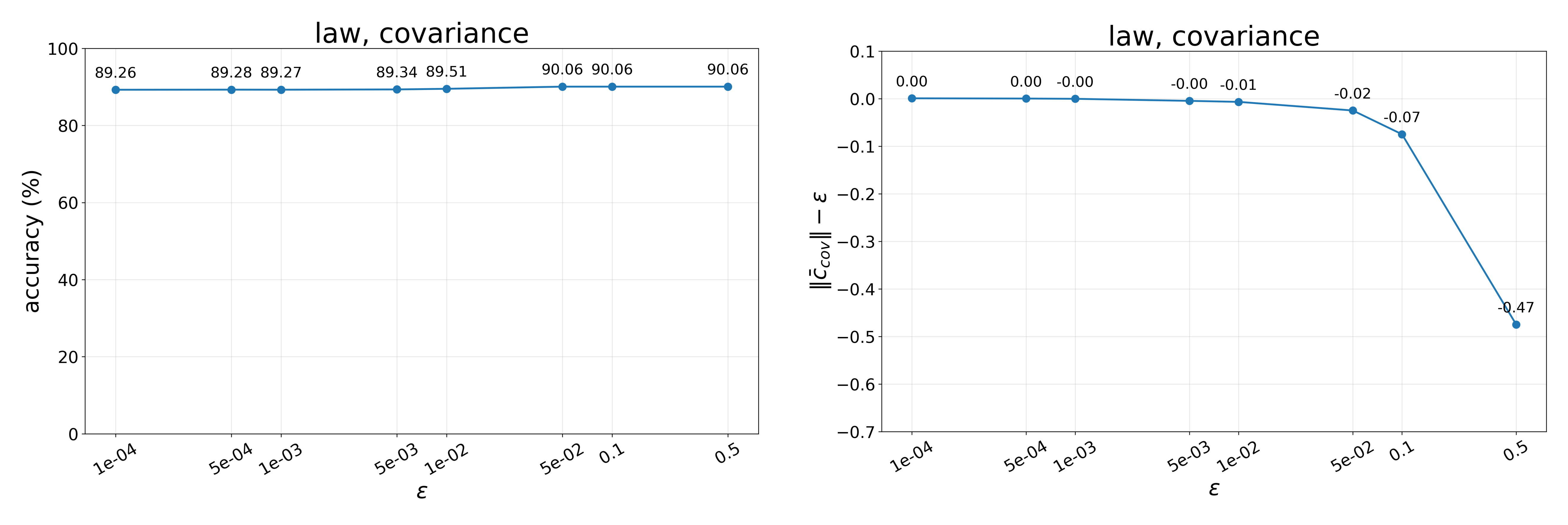}

    \caption{Training accuracy and constraint violation measures for smoothed-step, sigmoid, and covariance models on the Law dataset. The results indicate that fairness constraints can be satisfied without compromising prediction accuracy.}
    \label{fig:accuracy_measure}
\end{figure}

Figure~\ref{fig:accuracy_measure} confirms our prior claim that, through the use of hard constraints on nonconvex surrogates of unfairness measures, one is able to train prediction models that also offer high-accuracy predictions.  The results in this figure correspond to those in Figure~\ref{fig:delta_match}, i.e., with the models trained through scaled surrogate functions.  The prediction accuracies remain high, even as the value of $\delta$ or $\epsilon$ is adjusted to the point where the imposed hard constraints become active.

\subsection{Comparison of Constraint-Only-Based vs. Regularization-Only-Based Approaches}

Our second set of experiments demonstrates that enforcing hard constraints on unfairness-measure surrogates yields better prediction models and offers more precise control over unfairness measures.  We consider two approaches: A constraint-only-based approach that imposes hard constraints as defined by \eqref{eq:di} with no regularization (i.e., $\lambda = \infty$ in~\eqref{opt-prob}) versus a regularization-only-based approach that employs $\delta = 0$ and varying values of $\lambda$ in~\eqref{opt-prob}.

\begin{figure}[!ht]
    \centering
    \begin{subfigure}[b]{0.49\linewidth}
        \centering
        \includegraphics[width=\linewidth]{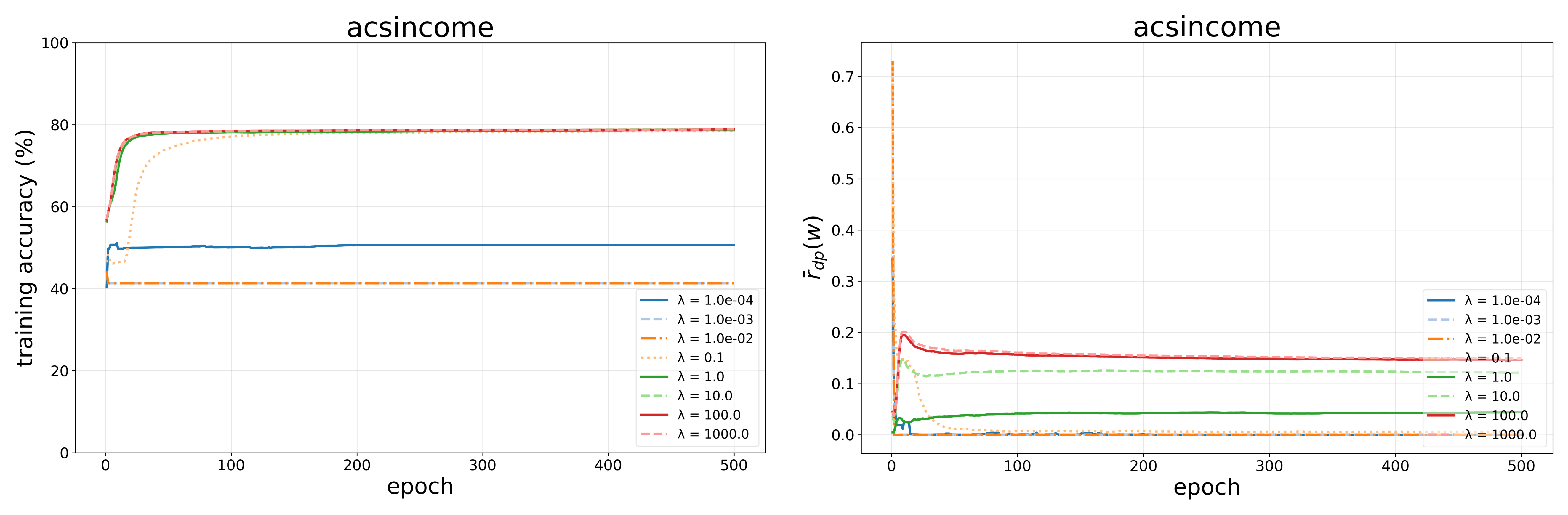}
        \caption{Constraint-only approach}
    \end{subfigure}
    \begin{subfigure}[b]{0.49\linewidth}
        \centering
        \includegraphics[width=\linewidth]{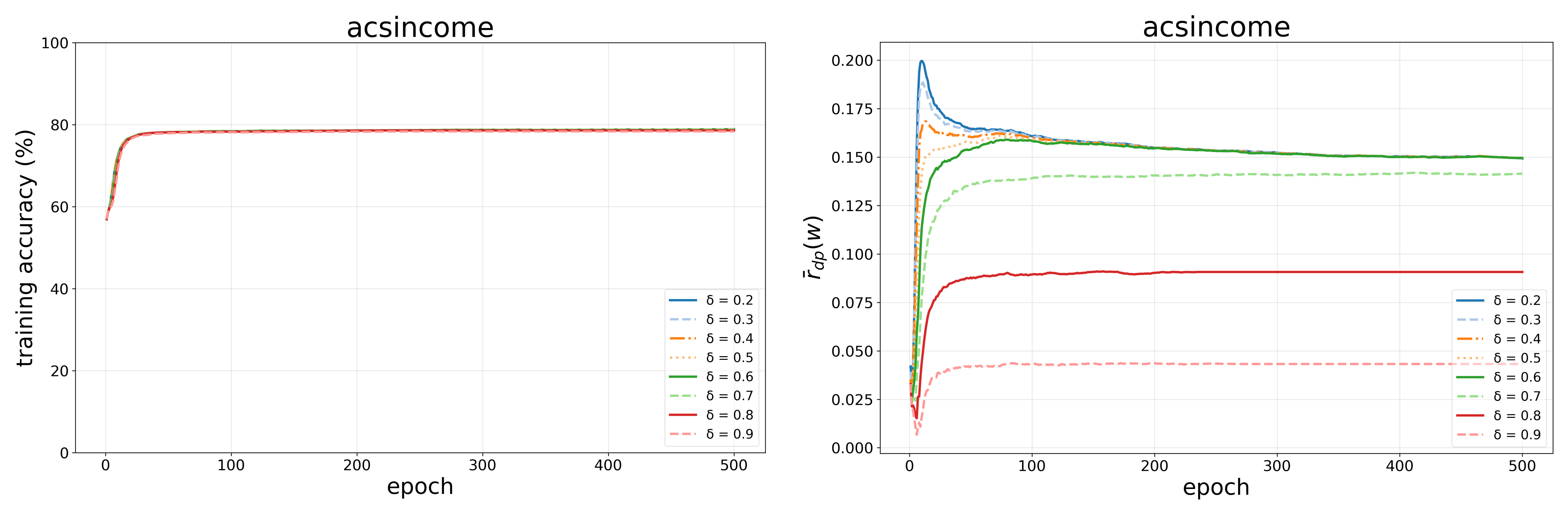}
        \caption{Regularization-only approach}
    \end{subfigure}

    \caption{Comparison between the constraint-only (left) and regularization-only (right) approaches for enforcing fairness using the smoothed-step surrogate on ACSIncome dataset. The constraint-only-based approach consistently meets the targeted threshold ($\delta$) with minimal impact on accuracy, while the regularization-only-based method exhibits unpredictable outcomes and greater accuracy loss, underscoring an advantage of using explicit constraints.}
    \label{fig:reg_const}
\end{figure}

Figure \ref{fig:reg_const} shows limitations of the regularization-only approach. First, the unfairness measure does not always improve monotonically with decreasing $\lambda$. This unexpected behavior is a consequence of formulating optimization problems that are increasingly difficult to solve as $\lambda$ decreases.  Second, the relationship between $\lambda$ and the resulting unfairness measure is nonlinear, thus requiring expensive tuning to reach a desired level of unfairness. Overall, the complicated relationship between the regularization parameter and unfairness measure makes it difficult to effectively use a regularization-only  approach.


In contrast, the constraint-only-based approach ensures monotonicity in the unfairness measure and allows a desired bound on unfairness (defined through group-specific probability ratios) to be enforced \emph{explicitly}.  Together, these properties allow the constraint-only-based approach to avoid requiring any significant hyperparameter tuning. It should also be noted that the constraint-only-based approach maintains high accuracy across many values of the unfairness thresholds, which is yet another advantage of it over the regularization-only-based approach.  


\subsection{Multiple Simultaneous Unfairness Measures}

The proposed framework can easily accommodate multiple unfairness constraints at the same time. To demonstrate this capability, we tested our framework when constraints on both disparate impact and equal impact were enforced simultaneously, each with varying thresholds. As shown in Table 2 in Appendix E, combining multiple unfairness constraints often leads to a reduction in accuracy that exceeds the reduction observed when only one of the constraints is enforced, especially as the thresholds become demanding. For example, for the ACSIncome dataset, enforcing a constraint only on EI yields an accuracy of 78.9\%, although that solution violates a constraint that one may place in DI.  Adding the additional DI constraint decreases the accuracy, but only slightly, to 78.5\%. Since our constraint-only-based approach consistently satisfies both  unfairness criteria when enforced as constraints, it highlights the practical value of our approach in real-world applications when multiple unfairness measures must be considered.

\section{Discussion and Conclusion}\label{sec.results}

We present a hard-constraint-based approach to fair supervised learning that accurately yields prediction models that meet prescribed limits on unfairness. Our method readily allows limits on multiple unfairness measures simultaneously, whereas attempting to invoke multiple measures through regularization terms can cause the training process to be extremely time-consuming and computationally expensive. A limitation of our approach is that there are slightly higher computational costs to train a model with hard constraints, although these additional costs are outweighed by the savings that our approach offers in terms of tuning costs.

\FloatBarrier
\bibliographystyle{plain}
\bibliography{reference}

\input{supplementary}

\end{document}

%% file: supplementary.tex
\section{Proof of Theorem 1}\label{app:proof}

For the purposes of the proof, let us simplify notation and define $\hat{y}_i(w) := \hat{y}(x_i,s_i,w)$ along with $t_i(w) := t(x_i,s_i,w)$ for all $i \in [N]$.  Let us also define the sample sets and their sizes given by
\begin{align*}
  {\cal N}_0^- := \{i \in [N] : s_i = 0 \wedge \hat{y}_i(w) = 0\}, &\ \ N_0^- := |{\cal N}_0^-|, \\
  {\cal N}_0^+ := \{i \in [N] : s_i = 0 \wedge \hat{y}_i(w) = 1\}, &\ \ N_0^+ := |{\cal N}_0^+|, \\
  {\cal N}_1^- := \{i \in [N] : s_i = 1 \wedge \hat{y}_i(w) = 0\}, &\ \ N_1^- := |{\cal N}_1^-|, \\ \text{and}\ \ 
  {\cal N}_1^+ := \{i \in [N] : s_i = 1 \wedge \hat{y}_i(w) = 1\}, &\ \ N_1^+ := |{\cal N}_1^+|.
\end{align*}

\proof{Proof.}
  Consider the shifted function $\varphi = 2(\phi - \tfrac12) = 2\phi - 1$, which under the conditions of the theorem is symmetric about the origin, meaning $\varphi(t) = -\varphi(-t)$ for all $t \in \mathbb{R}$.  One finds that
  \begin{align*}
    2c_{\textrm{dp}}(w) &= \frac{\sum_{\{i \in [N] : s_i = 1\}} 2\phi(t_i(w))}{N_1} - \frac{\sum_{\{i \in [N] : s_i = 0\}} 2\phi(t_i(w))}{N_0} \\
    &= \frac{\sum_{\{i \in [N] : s_i = 1\}} (\varphi(t_i(w)) + 1)}{N_1} - \frac{\sum_{\{i \in [N] : s_i = 0\}} (\varphi(t_i(w)) + 1)}{N_0} \\
    &= \frac{\sum_{\{i \in [N] : s_i = 1\}} \varphi(t_i(w))}{N_1} - \frac{\sum_{\{i \in [N] : s_i = 0\}} \varphi(t_i(w))}{N_0} \\
    &= \frac{\sum_{i \in {\cal N}_1^+} \varphi(|t_i(w)|) - \sum_{i \in {\cal N}_1^-} \varphi(|t_i(w)|)}{N_1^+ + N_1^-} - \frac{\sum_{i \in {\cal N}_0^+} \varphi(|t_i(w)|) - \sum_{i \in {\cal N}_0^-} \varphi(|t_i(w)|)}{N_0^+ + N_0^-}.
  \end{align*}
  Now the condition that $|c_{\textrm{dp}}(w)| \leq \epsilon$ implies that
  \begin{align*}
    \left|\frac{N_1^+ - N_1^-}{N_1^+ + N_1^-} \right. - \frac{N_0^+ - N_0^-}{N_0^+ + N_0^-} &+ \frac{N_1^- - N_1^+ + \sum_{i \in {\cal N}_1^+} \varphi(|t_i(w)|) - \sum_{i \in {\cal N}_1^-} \varphi(|t_i(w)|)}{N_1^+ + N_1^-} \\
    & + \left. \frac{N_0^+ - N_0^- - \sum_{i \in {\cal N}_0^+} \varphi(|t_i(w)|) + \sum_{i \in {\cal N}_0^-} \varphi(|t_i(w)|)}{N_0^+ + N_0^-} \right| \leq 2 \epsilon.
  \end{align*}
  This inequality has the form $|A + B| \leq 2\epsilon$, where by the triangle inequality one has that $|A + B| \geq |A| - |B|$, which in turn means that $|A| \leq |A + B| + |B| \leq 2\epsilon + |B|$.  Consequently, from above,
  \begin{align*}
    &\ \left| \frac{N_1^+ - N_1^-}{N_1^+ + N_1^-} - \frac{N_0^+ - N_0^-}{N_0^+ + N_0^-} \right| \\
    \leq&\ \left| \frac{N_1^- - N_1^+ + \sum_{i \in {\cal N}_1^+} \varphi(|t_i(w)|) - \sum_{i \in {\cal N}_1^-} \varphi(|t_i(w)|)}{N_1^+ + N_1^-} \right. \\
    &\ + \left. \frac{N_0^+ - N_0^- - \sum_{i \in {\cal N}_0^+} \varphi(|t_i(w)|) + \sum_{i \in {\cal N}_0^-} \varphi(|t_i(w)|)}{N_0^+ + N_0^-} \right| + 2\epsilon \\
    =&\ \left| \frac{\sum_{i \in {\cal N}_1^+} (\varphi(|t_i(w)|) - 1) - \sum_{i \in {\cal N}_1^-} (\varphi(|t_i(w)|) - 1)}{N_1^+ + N_1^-} \right. \\
    &\ + \left. \frac{- \sum_{i \in {\cal N}_0^+} (\varphi(|t_i(w)|) - 1) + \sum_{i \in {\cal N}_0^-} (\varphi(|t_i(w)|) - 1)}{N_0^+ + N_0^-} \right| + 2\epsilon.
  \end{align*}
  Now defining
  \begin{equation}
    S_1^+ = \sum_{i \in {\cal N}_1^+} (\varphi(|t_i(w)|) - 1)
  \end{equation}
  and similarly for $S_1^-$, $S_0^+$, and $S_0^-$, one finds from above that
  \begin{align*}
    &\ \left| \frac{S_1^+ - S_1^-}{N_1^+ + N_1^-} - \frac{S_0^+ - S_0^-}{N_0^+ + N_0^-} \right| + 2\epsilon \\
    \geq&\ \left| \frac{N_1^+ - N_1^-}{N_1^+ + N_1^-} - \frac{N_0^+ - N_0^-}{N_0^+ + N_0^-} \right| \\
    \geq&\ \left| \frac{N_1^+}{N_1^+ + N_1^-} - \left(1 - \frac{N_1^+}{N_1^+ + N_1^-}\right) - \frac{N_0^+}{N_0^+ + N_0^-} + \left(1 - \frac{N_0^+}{N_0^+ + N_0^-}\right) \right| \\
    =&\ 2\left| \frac{N_1^+}{N_1^+ + N_1^-} - \frac{N_0^+}{N_0^+ + N_0^-} \right|,
  \end{align*}
  which in turn implies that
  \begin{align*}
    |\bar{c}_{\textrm{dp}}(w)| = &\left|\frac{N_1^+}{N_1^+ + N_1^-} - \frac{N_0^+}{N_0^+ + N_0^-}\right| \leq \frac{1}{2} \left(\left|\frac{S_1^+ - S_1^-}{N_1^+ + N_1^-} - \frac{S_0^+ - S_0^-}{N_0^+ + N_0^-}\right| + 2\epsilon\right).
  \end{align*}
  Under the conditions of the theorem, one has for all $i \in [N]$ that $\phi(t_i(w)) \in [0,\gamma] \cup [1-\gamma,1]$, which in turn means that $\varphi(|t_i(w)|) \in [1-2\gamma,1] \subseteq [1-2\gamma,1+2\gamma]$, so $S_1^+ \in [-2\gamma N_1^+, 2\gamma N_1^+]$ and similarly for $S_1^-$, $S_0^+$, and $S_0^-$.  Thus, $\frac{S_1^+ - S_1^-}{N_1^+ + N_1^-} \in [-2\gamma,2\gamma]$ and $\frac{S_0^+ - S_0^-}{N_0^+ + N_0^-} \in [-2\gamma,2\gamma]$, which in turn shows
  \begin{equation*}
    \left|\frac{S_1^+ - S_1^-}{N_1^+ + N_1^-} - \frac{S_0^+ - S_0^-}{N_0^+ + N_0^-}\right| \in [0,2 \times 2\gamma].
  \end{equation*}
  With the prior conclusion, one obtains that $|\bar{c}_{\textrm{dp}}(w)| \leq 2\gamma + \epsilon$, as desired.
\endproof

For any $\epsilon > 0$, we can choose a sufficiently large scaling factor $\alpha$ such that the scaled surrogate constraint $|c_{dp}(w, \alpha)| \leq \epsilon$ implies that the true constraint is satisfied approximately: $|\bar{c}{dp}(w)| \leq \epsilon + 2\gamma_\alpha$, where $\gamma_\alpha \to 0$ as $\alpha \to \infty$. 

Note: the bound in theorem 1 of the main paper should be $|\bar{c}_{\textrm{dp}}(w)| \leq 2\gamma + \epsilon$.

\section{Formal Statement and Proof of Theorem 3}\label{sec.unbiased}

Let the features be defined by a pair of random variables $(X,S)$, which are in turn defined by a probability measure $(\Omega,{\cal F},\mathbb{P})$.  Here, $S$ represents the sensitive feature, and $S(\omega) \in \{0,1\}$ for all $\omega \in \Omega$.  Subsequently, let $\{(X_i,S_i)\}_{i=1}^N$ be a set of $N$ random-variable pairs, where for all $i \in [N]$ the pair $(X_i,S_i)$ has the same distribution as $(X,S)$.  The set of possible outcomes of $\{(X_i,S_i)\}_{i=1}^N$ is $\Omega \times \cdots \times \Omega = \prod_{i=1}^N \Omega$, and the corresponding $\sigma$-algebra and probability measure for $\{(X_i,S_i)\}_{i=1}^N$, call it $\mathbb{P}_N$, can be derived from $({\cal F},\mathbb{P})$.

In general, a realization of $\{(X_i,S_i)\}_{i=1}^N$, call it $\{(x_i,s_i)\}_{i=1}^N$, can have $0 \leq \sum_{i=1}^n s_i \leq N$.  (For example, if $S$ indicates female or male, then a random sample of $N$ data points could have any numbers of females or males.)  It would be problematic to work with all such realizations when aiming to ensure an unbiased estimate of an unfairness measure; e.g., if a sample contains no males or no females, then the unfairness-measure estimate is not well defined.  Instead, we want to work with the conditional distribution of $\{(X_i,S_i)\}_{i=1}^N$ subject to the condition that any realization has $N_0 \in \{1,\dots,N\}$ values with $s_i = 0$ and $N_1 = N - N_0$ values with $s_i = 1$.  Let this conditional distribution have measure $\mathbb{P}_{N_0,N_1}$.  Let $\mathbb{E}_{N_0,N_1}$ denote expectation taken with respect to $\mathbb{P}_{N_0,N_1}$.

At the same time, we can define the conditional distribution of $X$ subject to $S = 0$ and the conditional distribution of $X$ subject to $S = 1$.  Let the corresponding probability measures be $\mathbb{P}_0$ and $\mathbb{P}_1$, respectively.  Let $\mathbb{E}_0$ and $\mathbb{E}_1$ denote expectation taken with respect to $\mathbb{P}_0$ and $\mathbb{P}_1$, respectively.

Given $w$, the values in which we are interested are the real number
\begin{equation*}
  c(w) = \mathbb{E}_0 [ \phi(t(X, 0, w)) ] - \mathbb{E}_1 [ \phi(t(X, 1, w)) ]
\end{equation*}
and the random variable
\begin{equation*}
  C(w) = \frac{1}{N_0} \sum_{\{i : S_i = 0\}} \phi(t(X_i, 0, w)) - \frac{1}{N_1} \sum_{\{i : S_i = 1\}} \phi(t(X_i, 1, w)).
\end{equation*}

\textbf{Theorem 3. (Formal)} \textit{For any w, the random variable $C(w)$ is an unbiased estimator of $c(w)$. That is, $\mathbb{E}_{N_0,N_1}[C(w)] = c(w)$ for all $w$.}

\proof{Proof.}
We start by taking the expectation of $C(w)$
\[
\mathbb{E}_{N_0,N_1}[C(w)] = \mathbb{E}_{N_0,N_1} \left[\frac{1}{N_0} \sum_{\{i : S_i = 0\}} \phi(t(X_i, 0, w)) - \frac{1}{N_1} \sum_{\{i : S_i = 1\}} \phi(t(X_i, 1, w))\right]
\]
Using the linearity of expectation
\[
\mathbb{E}_{N_0,N_1}[C(w)] = \mathbb{E}_{N_0,N_1}\left[\frac{1}{N_0} \sum_{\{i : S_i = 0\}} \phi(t(X_i, 0, w))\right] - \mathbb{E}_{N_0,N_1}\left[\frac{1}{N_1} \sum_{\{i : S_i = 1\}} \phi(t(X_i, 1, w))\right]
\]
Using the property that the expectation of a sum is the sum of expectations
\[
\mathbb{E}_{N_0,N_1}[C(w)] = \frac{1}{N_0} \sum_{\{i : S_i = 0\}} \mathbb{E}_0\left[\phi(t(X, 0, w))\right] - \frac{1}{N_1} \sum_{\{i : S_i = 1\}} \mathbb{E}_1\left[\phi(t(X, 1, w))\right]
\]
Since the samples are i.i.d., the expectation of each term in the sums is the same
\[
\mathbb{E}_{N_0,N_1}[C(w)] = \mathbb{E}_0\left[\phi(t(X, 0, w))\right] - \mathbb{E}_1\left[\phi(t(X, 1, w))\right]
\]
By comparing the above result with the definition of $c(w)$, we see that
\[
\mathbb{E}_{N_0,N_1}[C(w)] = c(w),
\]
as claimed.
\endproof

\section{Overview of Datasets}\label{app:datasets}

An overview of the datasets employed in our numerical experiments is provided in Table~\ref{tab.overview}.  Along with basic information including the number of data points $N$, numbers of data points with sensitive feature value equal to 1 or 0, etc., the table includes some statistics that are relevant for measuring how innately fair or unfair is the dataset.  These are motivated in the following paragraphs.  In each of the descriptions below, the measure $\mathbb{P}$ corresponds to the empirical distribution of the dataset.

\paragraph{Demographic Parity Violation with Rote Learning (DPVRL).}

If a prediction model were to predict the true labels with 100\% accuracy, i.e., $\hat{Y} = Y$ over the entire dataset, which corresponds to rote learning of the dataset, then the violation of demographic parity is given by
\begin{equation*}
  | \mathbb{P}(Y = 1 | S = 1) - \mathbb{P}(Y = 1 | S = 0) |.
\end{equation*}
This measure quantifies the absolute difference in the likelihood of a positive outcome $Y = 1$ between one of the groups ($S = 1$) and the other group ($S = 0$) based on the sensitive feature.

\paragraph{Demographic Representation Imbalance} Violation with Rote Learning (DRIVRL).

If a prediction model were to predict the true labels with 100\% accuracy, then the violation of Demographic Representation Imbalance is given by
\begin{equation*}
  | \mathbb{P}(S = 1 | Y = 1) - \mathbb{P}(S = 0 | Y = 1) |.
\end{equation*}
This measure assesses the disparity in true positive rates between the protected and unprotected groups, focusing on individuals who actually received a positive outcome ($Y = 1$).

\paragraph{Disparate Impact with Rote Learning (DIRL)}

If a prediction model were to predict the true labels with 100\% accuracy, then the disparate impact---namely, the largest value of $\delta$ such that both of the inequalities in Equation (5) of the main paper would be satisfied---is given by
\begin{equation*}
  \min\left\{ \frac{\mathbb{P}(Y = 1 | S = 0)}{\mathbb{P}(Y = 1 | S = 1)}, \frac{\mathbb{P}(Y = 1 | S = 1)}{\mathbb{P}(Y = 1 | S = 0)} \right\}.
\end{equation*}
This measures the relative disparity in the likelihood of a positive outcome between the protected and unprotected groups, providing insight into potential biases in the model's predictions.  For the datasets shown in Table~\ref{tab.overview}, this measure provides insight into why our constraint on disparate impact was not active until $\delta \approx 0.78$ for the Law dataset and was active with $\delta \approx 0.36$ for the Adult dataset.

\paragraph{All-Zero Prediction and All-One Prediction Accuracies}

A potential concern in the pursuit of a fair prediction model is that model training would result in consistent predictions for all individuals regardless of their feature data.  This corresponds to no learning, and simply to predicting the same outcome for all individuals.  To determine whether such predictions could be desirable from the viewpoint of achieving near-optimal accuracy, one can compute
\begin{equation*}
  \frac{\mathbb{P}(Y = 0)}{\mathbb{P}(Y = 0) + \mathbb{P}(Y = 1)}\ \ \text{and}\ \ \frac{\mathbb{P}(Y = 1)}{\mathbb{P}(Y = 0) + \mathbb{P}(Y = 1)}
\end{equation*}
For the datasets shown in Table~\ref{tab.overview}, one finds that the optimal accuracy in terms of a learned prediction model should be at least $89.1\%$ for the Law dataset and at least $75.6\%$ for the Adult dataset.
{\small
\begin{table}[ht]
  \centering
  \caption{Statistics for the Dutch, Law, and ACSIncome datasets.}
  \label{tab.overview}
  \begin{tabular}{lccc}
    \toprule
    \textbf{Metric} & \textbf{Dutch} & \textbf{Law} & \textbf{ACSIncome} \\
    \midrule
    $N$ & 60,420 & 16,638 & 50,000 \\
    \# $S = 1$ & 30,147 & 13,995 & 26,441 \\
    \# $S = 0$ & 30,273 & 2,643 & 23,559 \\
    \# $Y = 1$ & 31,657 & 14,826 & 20,661 \\
    \# $Y = 0$ & 28,763 & 1,812 & 29,339 \\
    \# $Y = 1 \wedge S = 1$ & 11,287 & 12,915 & 12,350 \\
    \# $Y = 1 \wedge S = 0$ & 20,370 & 1,911 & 8,311 \\
    \# $Y = 0 \wedge S = 1$ & 18,860 & 1,080 & 14,091 \\
    \# $Y = 0 \wedge S = 0$ & 9,903 & 732 & 15,248 \\
    \midrule
    $\mathbb{P}(Y = 1 | S = 1)$ & 0.3744 & 0.923 & 0.4672 \\
    $\mathbb{P}(Y = 1 | S = 0)$ & 0.6730 & 0.723 & 0.3528 \\
    $\mathbb{P}(S = 1 | Y = 1)$ & 0.3565 & 0.871 & 0.5975 \\
    $\mathbb{P}(S = 0 | Y = 1)$ & 0.6435 & 0.129 & 0.4025 \\
    \midrule
    DPVRL & 0.2986 & 0.200 & 0.1144 \\
    DRIVRL & 0.2870 & 0.743 & 0.1950 \\
    DIRL & 0.5564 & 0.784 & 0.7553 \\
    \midrule
    All-Zero Pred Acc & 0.4758 & 0.109 & 0.5868 \\
    All-One Pred Acc & 0.5242 & 0.891 & 0.4132 \\
    \bottomrule
  \end{tabular}
\end{table}}

\section{SQP Algorithm and Learning Rate Adjustment Strategy}\label{app:algorithms}

For the purposes of describing the SQP algorithm (based on that in \cite{curtis2023fair}) that we employ for model training for our experiments in the paper, let us write the optimization problem of interest in the following form, where we consider the case of having two constraint functions:
    
\begin{align}\label{training_obj}
\min_w\ \ \sum_{i \in [N]} \ell({\cal N}(x_i,s_i,w),y_i) + \tfrac{1}{\lambda} r(w)\\ \text{subject to}\ \ \begin{bmatrix} l_1 \\ l_2 \end{bmatrix} \leq \begin{bmatrix} c_1(w) \\ c_2(w) \end{bmatrix} \leq \begin{bmatrix} u_1 \\ u_2 \end{bmatrix}.
\end{align}
The algorithm that we state is readily extended to the case of more than 2 constraint functions.

Each iteration of the SQP algorithm involves solving a quadratic optimization (QP) subproblem to compute a step.  In each iteration $k \in \mathbb{N}$, in which the current iterate is $w_k$, this QP has the form
{\small
\begin{equation}
  \begin{aligned}\label{eq: feasibility}
    & \min_d  \ g(w_k)^Td + \tfrac{1}{2} d^T H_k d \\
    &\text{s.t.} \\
    & 
    J(w_k) d := \begin{bmatrix} -\nabla c_1(w_k)^T \\ \nabla c_1(w_k)^T \\ -\nabla c_2(w_k)^T \\ \nabla c_2(w_k)^T \end{bmatrix} d \leq - \begin{bmatrix} l_1 - c_1(w_k) \\ c_1(w_k) - u_1 \\ l_2 - c_2(w_k) \\ c_2(w_k) - u_2 \end{bmatrix} =: -r(w_k).
  \end{aligned}
\end{equation}}

where $g(w_k)$ is a gradient or stochastic gradient of the objective function of \eqref{training_obj} at $w_k$ and $H_k$ is a positive definite matrix.  Since $H_k$ is positive definite and the feasible region of this QP is convex, it follows that the QP has a unique optimal solution.  Moreover, as is well known, if it is known which of the inequality constraints are satisfied at equality at the optimal solution of the QP, then the solution can be obtained by solving a symmetric indefinite linear system of the form
\begin{equation}\label{eq: kkt}
  \begin{bmatrix} H_k & J_{\cal A}(w_k) \\ J_{\cal A}(w_k)^T & 0 \end{bmatrix} \begin{bmatrix} d \\ y \end{bmatrix} = -\begin{bmatrix} g \\ r_{\cal A}(w_k) \end{bmatrix}.
\end{equation}
Here, $J_{\cal A}$ and $r_{\cal A}$ are the rows of $J$ and $r$ (see \eqref{eq: feasibility}) corresponding to the indices in ${\cal A} \subseteq \{1,2,3,4\}$.

Our SQP algorithm, stated as Algorithm~\ref{alg:sqp} below, specifies a procedure for solving \eqref{eq: feasibility} in each iteration by considering, when necessary, different choices for the optimal active set ${\cal A}$.  We note that for computational efficiency any matrix of the form in \eqref{eq: kkt} only needs to be factored once in order to solve any number of linear systems of the form \eqref{eq: kkt} in the iteration.


\begin{algorithm}[!ht]
  \caption{Stochastic Sequential Quadratic Programming (SQP)}
  \label{alg:sqp}
  \begin{algorithmic}[1]
    \State \textbf{Require:} initial iterate $w_0$, prescribed learning-rate sequence $\{\beta_k\}$, initial merit parameter $\tau_{-1}$
    \For{$k = 0, 1, 2, ...$}
      \State Choose $H_k$ (our implementation uses an Adagrad-type scheme)
      \State Compute $d_{mm}$ as the unconstrained minimizer of \eqref{eq: feasibility}, i.e., solve \eqref{eq: kkt} with ${\cal A} = \emptyset$
      \If{$d_{mm}$ is feasible for \eqref{eq: feasibility}}
        \State Set $d_k \gets (d_{mm})$
      \Else
        \State Compute $d_{lm}$ as the minimizer of \eqref{eq: feasibility} for ${\cal A} = \{1\}$
        \State Compute $d_{um}$ as the minimizer of \eqref{eq: feasibility} for ${\cal A} = \{2\}$
        \State Compute $d_{ml}$ as the minimizer of \eqref{eq: feasibility} for ${\cal A} = \{3\}$
        \State Compute $d_{mu}$ as the minimizer of \eqref{eq: feasibility} for ${\cal A} = \{4\}$
        \If{any of these candidates is feasible for \eqref{eq: feasibility}}
          \State Set $d_k = \underset{d \in \{d_{lm}, d_{um}, d_{ml}, d_{mu}\}}{\operatorname{argmin}} g(w_k)^Td + \tfrac{1}{2} d^T H_k d$
        \Else
          \State Compute $d_{ll}$ as the minimizer of \eqref{eq: feasibility} for ${\cal A} = \{1,3\}$
          \State Compute $d_{lu}$ as the minimizer of \eqref{eq: feasibility} for ${\cal A} = \{1,4\}$
          \State Compute $d_{ul}$ as the minimizer of \eqref{eq: feasibility} for ${\cal A} = \{2,3\}$
          \State Compute $d_{uu}$ as the minimizer of \eqref{eq: feasibility} for ${\cal A} = \{2,4\}$
          \State Set $d_k = \underset{d \in \{d_{ll}, d_{lu}, d_{lu}, d_{uu}\}}{\operatorname{argmin}} g(w_k)^Td + \tfrac{1}{2} d^T H_k d$ 
        \EndIf
      \EndIf
      \State Set $\tau_k$ based on $d_k$; see \cite{curtis2023fair}
      \State Set $\alpha_k$ based on $\beta_k$ and $\tau_k$; see \cite{curtis2023fair}
      \State Set $w_{k+1} \gets w_k + \alpha_k d_k$
      \State \textbf{Apply Learning Rate Adjustment Strategy (Algorithm 2)}
      \State $k \gets k + 1$
  \EndFor
\end{algorithmic}
\end{algorithm}

\begin{algorithm}[!ht]
  \caption{Learning Rate Adjustment Strategy (based on \cite{zhang2004nonmonotone})}
  \label{alg:lr_adjustment}
  \begin{algorithmic}[1]
    \Require 
    \State $\tau_k$ \Comment{merit parameter}
    \State $\alpha_{\text{min}} \gets 10^{-7}$ \Comment{minimum learning rate}
    \State $k_{\text{min}} \gets 200$ \Comment{minimum iterations}
    \State $\Delta k \gets 5$ \Comment{adjustment interval}
    \State $\gamma \gets 10$ \Comment{reduction factor}
    \State $\eta \gets 0.85$ \Comment{initial weight for convex combination}
    \State $\delta \gets 0$ \Comment{iterations since last adjustment}
    \State $n_c$ \Comment{number of constraints}
    \State $f(w_k)$ \Comment{objective value at iteration $k$ (see \eqref{training_obj})}
    \State $r(w_k)$ \Comment{constraint value at iteration $k$ (see \eqref{eq: feasibility})}
    \State $\phi_k \gets \tau f(w_k) + \|r(w_k)\|_1$ \Comment{compute merit value}
    \If{$k = 0$}
      \State Set $\bar{\phi}_k \gets \phi_k$ \Comment{initialize moving average}
    \Else
      \State Set $\bar{\phi}_k \gets \eta \phi_k + (1 - \eta) \bar{\phi}_{k-1}$ \Comment{update moving average}
    \EndIf
    \If{$k \geq k_{\text{min}}$ and $\alpha_k \geq \alpha_{\text{min}}$}
    \If{$\bar{\phi}_k \leq \phi_k$}
        \If{$\delta_k \geq \Delta k$}
            \State $\alpha_{k+1} \gets \alpha_k/\gamma$
            \State $\delta_k \gets 0$
        \EndIf
    \EndIf
    \State $\delta_k \gets \delta_k + 1$
\EndIf
\end{algorithmic}
\end{algorithm}

\section{Additional Experimental Results}
\label{app:plots}

This section provides supplementary experimental results that complement the findings presented in the experiments section. We first report the compute times for training of each model and dataset at Table ~\ref{tab:comp_budget}. Following this, we present additional plots.

\begin{figure}[ht]
    \centering
    \begin{subfigure}[b]{0.49\textwidth}
        \centering
        \includegraphics[width=\textwidth]{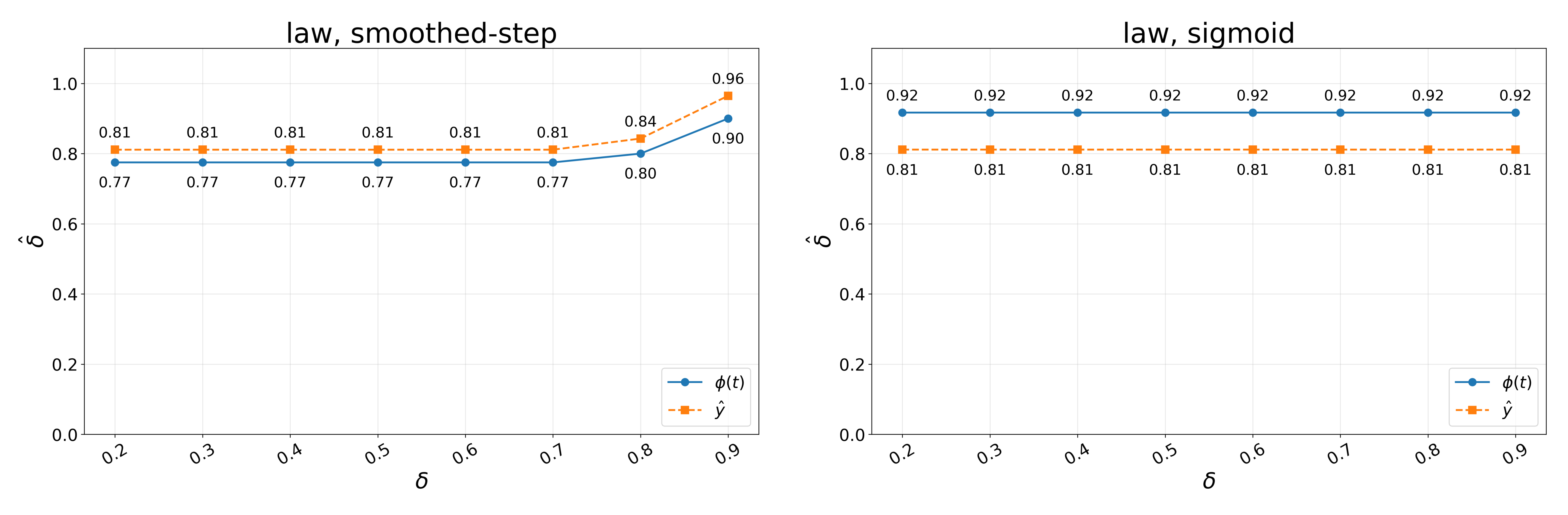}
    \end{subfigure}%
    \hfill
    \begin{subfigure}[b]{0.49\textwidth}
        \centering
        \includegraphics[width=\textwidth]{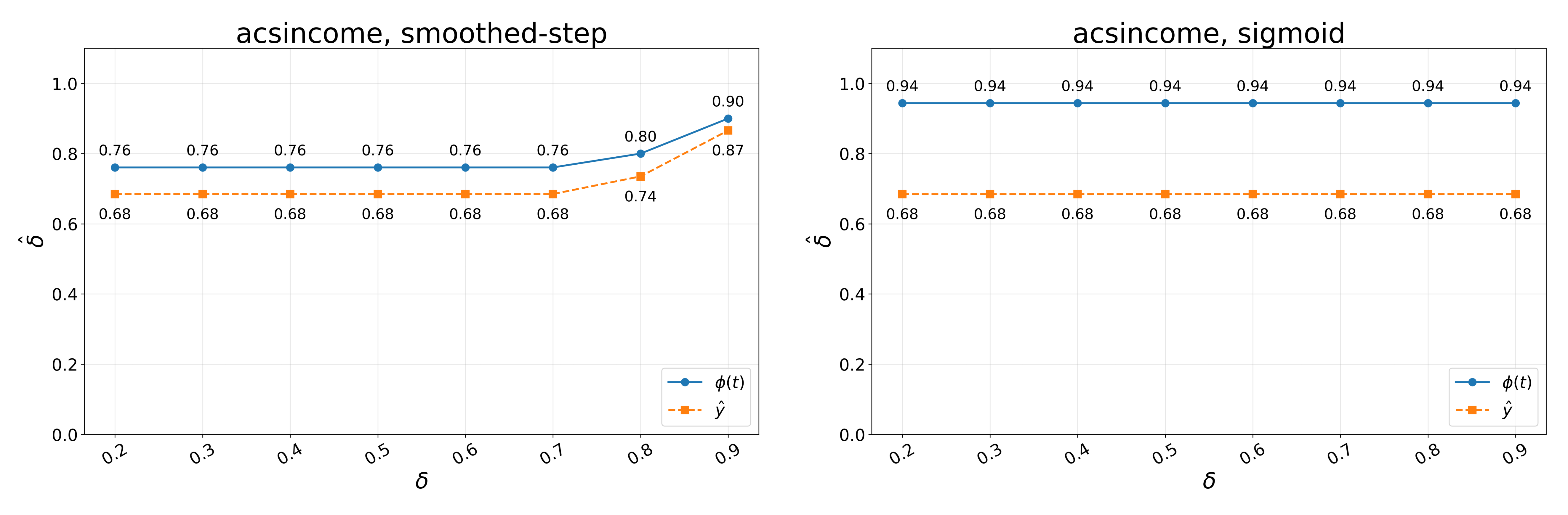}
    \end{subfigure}
    \caption{Levels of disparate impact actually achieved ($\hat\delta$) when prediction models are trained with constraints as in Equation (5) of the main paper for varying values of $\delta$ \emph{without} scaling of the surrogate functions.  The surrogate approximations not being tight causes large gaps between the levels of disparate impact desired and the levels actually achieved.  The graphs indicated by $\phi(t)$ show the values such that $c_{\textrm{di}}(w) \leq 0$, whereas the graphs indicated by $\hat{y}$ show the values such that $\bar{c}_{\textrm{di}}(w) \leq 0$.}
    \label{fig:delta_gap}
\end{figure}

\begin{figure}[ht]
    \centering
    \begin{subfigure}[b]{0.7\textwidth}
        \centering
        \includegraphics[width=\textwidth]{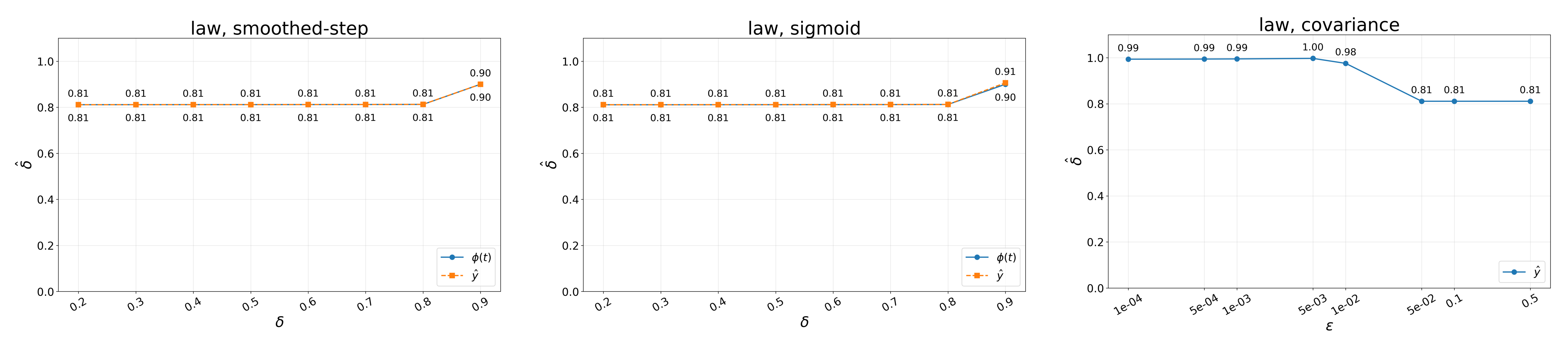}
    \end{subfigure}
    \vspace{0.1cm} 
    \begin{subfigure}[b]{0.7\textwidth}
        \centering
        \includegraphics[width=\textwidth]{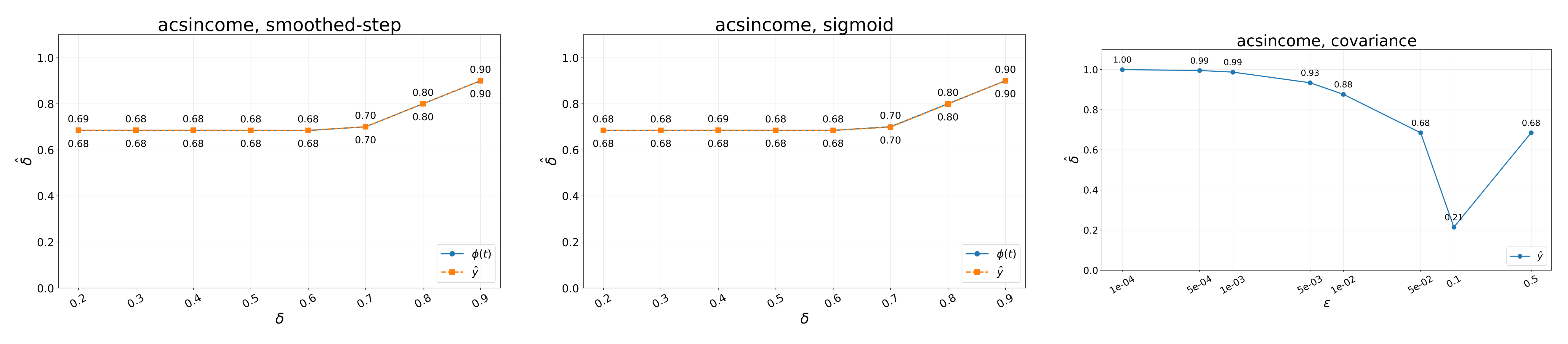}
    \end{subfigure}
    \caption{Levels of disparate impact actually achieved ($\hat\delta$) when prediction models are trained with constraints as in Equation (5) of the main paper for varying values of $\delta$ \emph{with} scaling of the surrogate functions.  These results should be contrasted with those in Figure~\ref{fig:delta_gap}.  In particular, it should be observed that scaling the surrogate functions leads to much tighter correspondence between $\delta$ and $\hat\delta$ when the constraints are tight.  Also, the plots on the right are the levels of disparate impact actually achieved when a constraint on a covariance surrogate (less than or equal to $\epsilon$) is imposed.}
    \label{fig:delta_match}
\end{figure}

\begin{figure}[ht]
    \centering
    \begin{subfigure}[b]{0.49\textwidth}
        \centering
        \includegraphics[width=\textwidth]{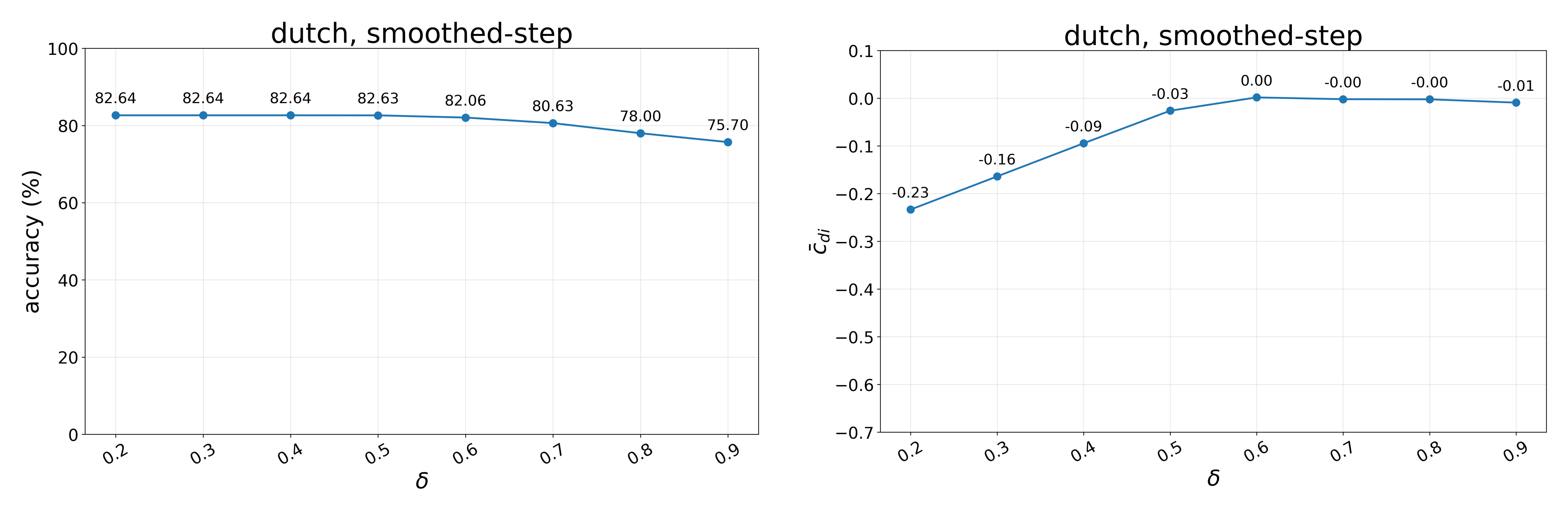}
    \end{subfigure}
    \begin{subfigure}[b]{0.49\textwidth}
        \centering
        \includegraphics[width=\textwidth]{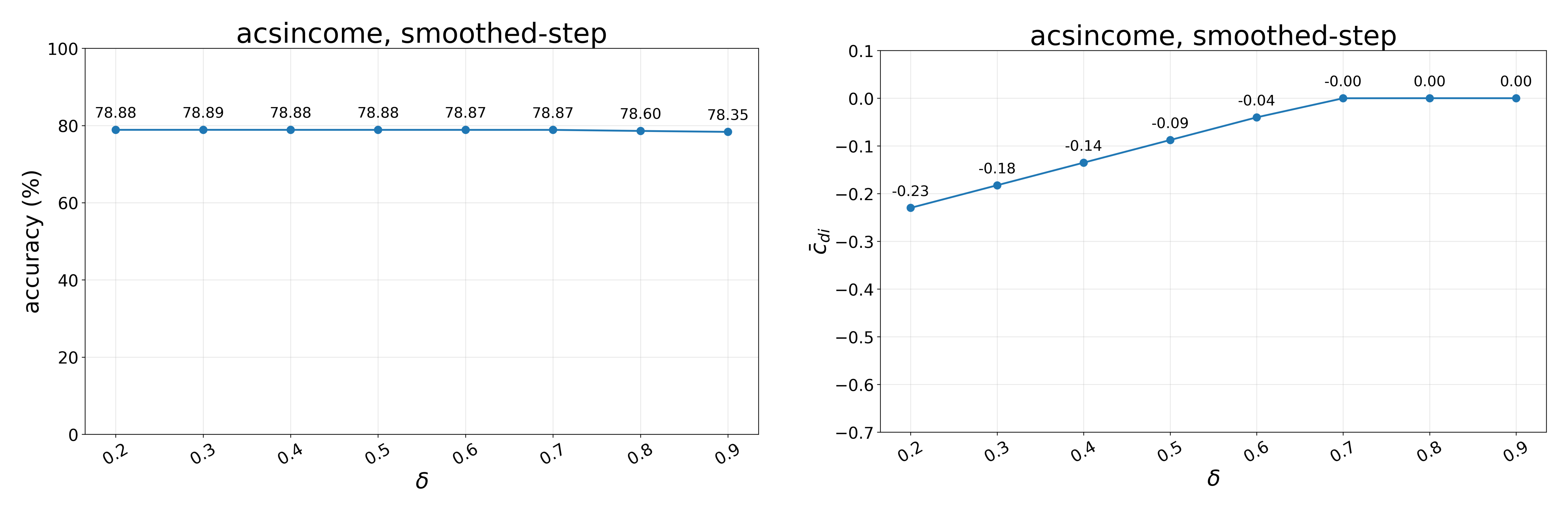}
    \end{subfigure}
    \vspace{0.2cm}
    \begin{subfigure}[b]{0.49\textwidth}
        \centering
        \includegraphics[width=\textwidth]{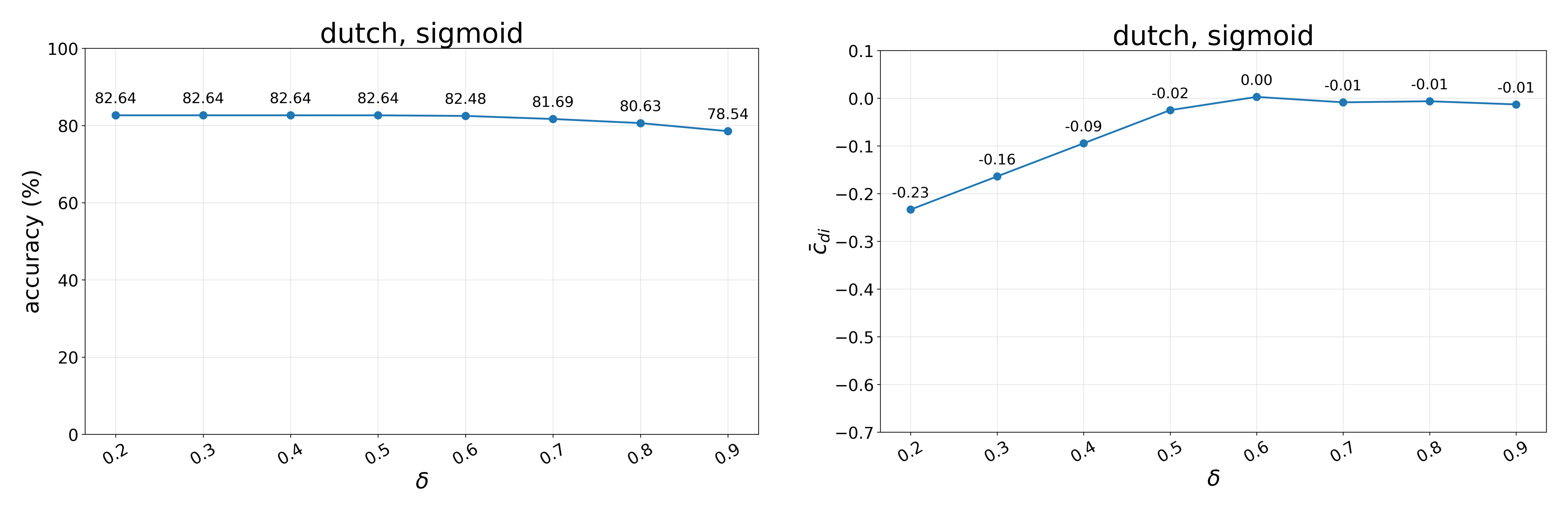}
    \end{subfigure}
    \begin{subfigure}[b]{0.49\textwidth}
        \centering
        \includegraphics[width=\textwidth]{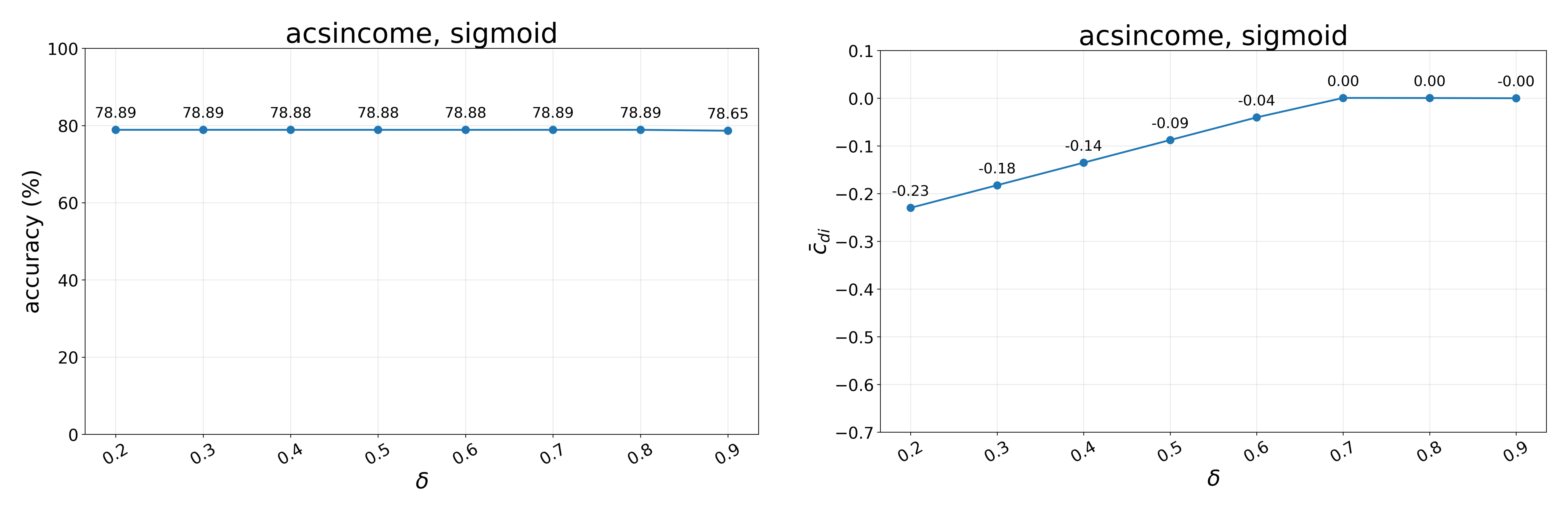}
    \end{subfigure}

    \vspace{0.2cm}
    
    \begin{subfigure}[b]{0.49\textwidth}
        \centering
        \includegraphics[width=\textwidth]{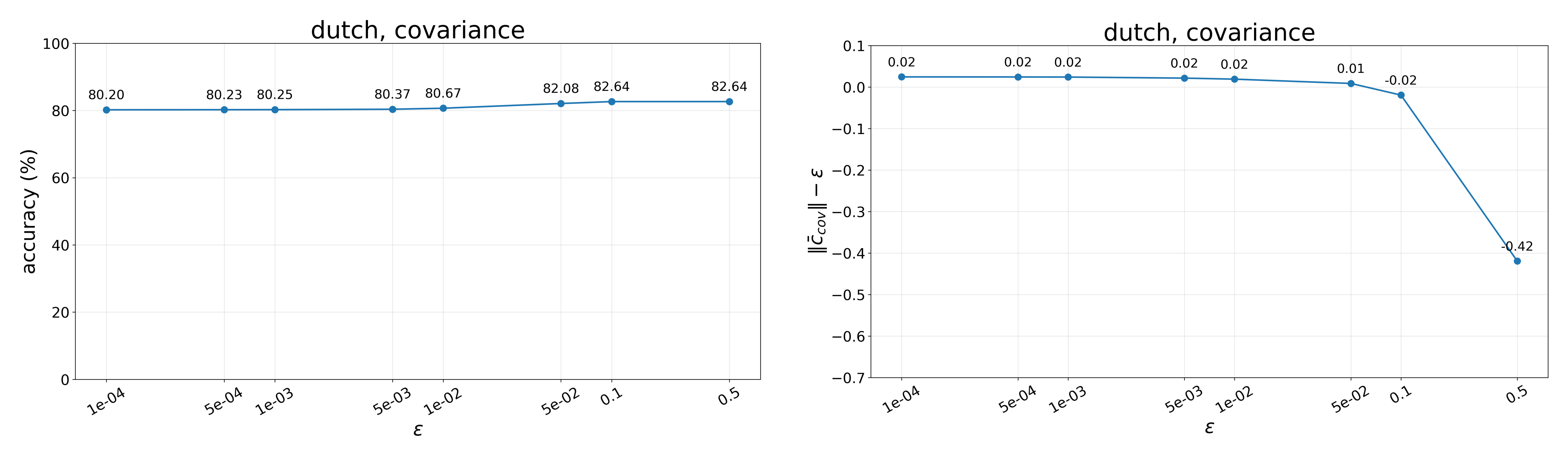}
    \end{subfigure}
    \begin{subfigure}[b]{0.49\textwidth}
        \centering
        \includegraphics[width=\textwidth]{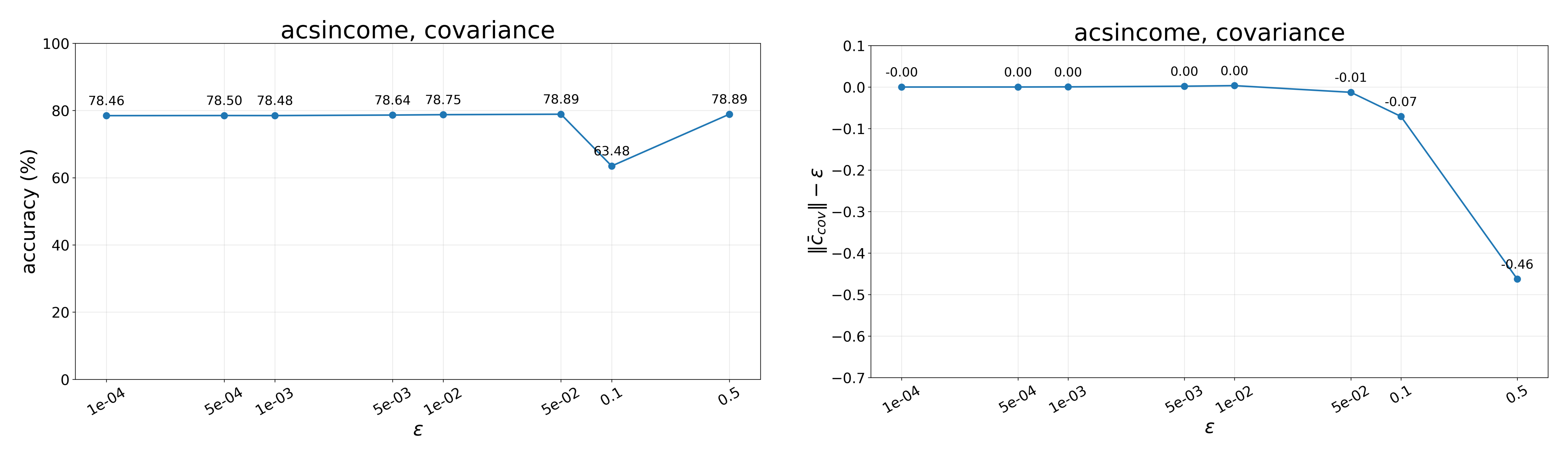}
    \end{subfigure}

    \caption{Training accuracy and constraint violation measures for smoothed-step, sigmoid, and covariance models on the Dutch and ACSIncome datasets. The plots show training accuracy alongside constraint violation metrics for different surrogate functions. The results indicate that fairness constraints can be satisfied without compromising prediction accuracy.}
    \label{fig:accuracy_measure}
\end{figure}

\begin{figure}[ht]
\centering
\begin{subfigure}[b]{0.49\textwidth}
\centering
\includegraphics[width=\textwidth]{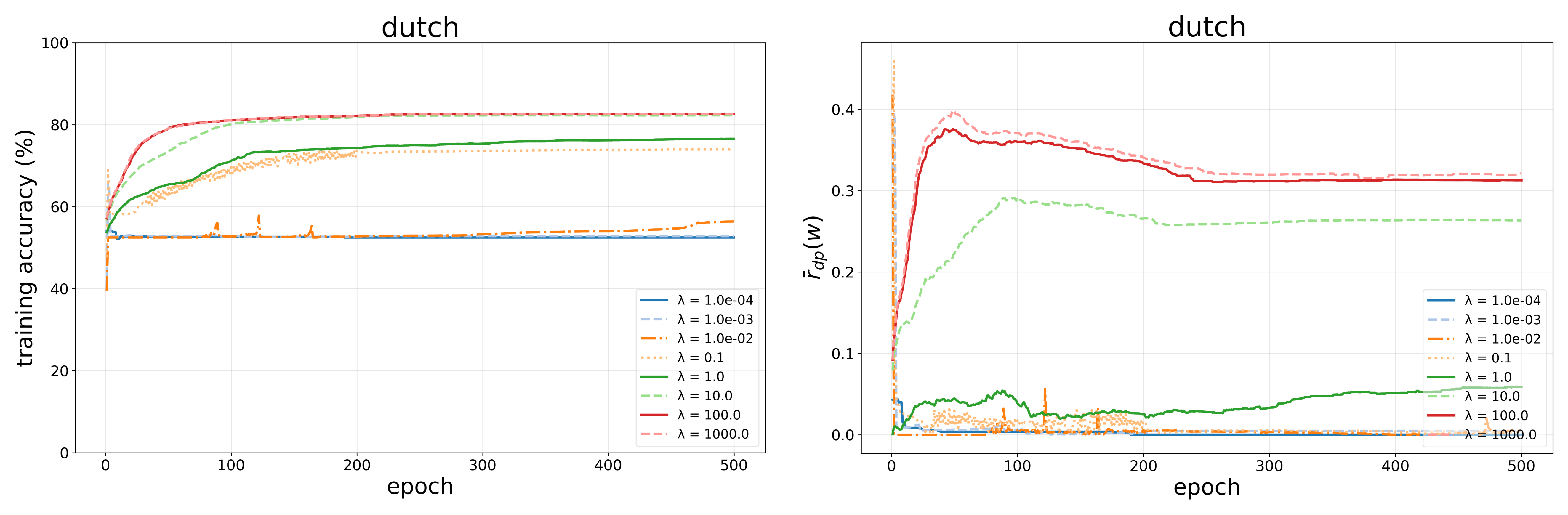}
\end{subfigure}%
\hfill
\begin{subfigure}[b]{0.49\textwidth}
\centering
\includegraphics[width=\textwidth]{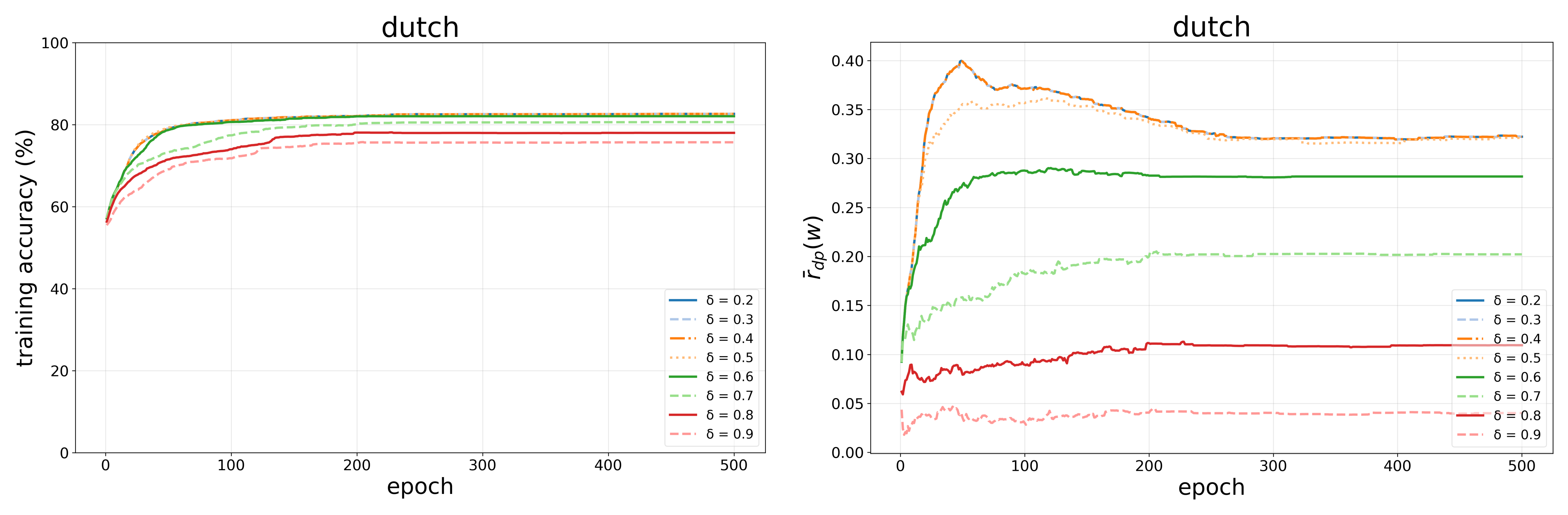}
\end{subfigure}

\vspace{0.3cm}

\begin{subfigure}[b]{0.49\textwidth}
\centering
\includegraphics[width=\textwidth]{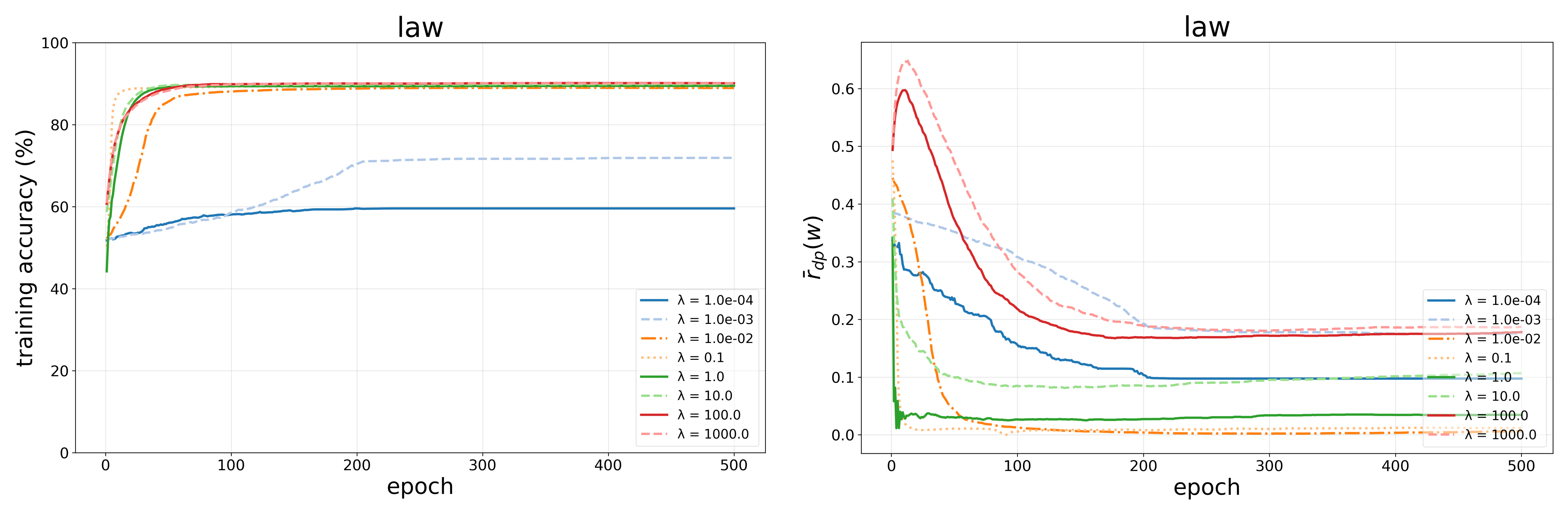}
\end{subfigure}%
\hfill
\begin{subfigure}[b]{0.49\textwidth}
\centering
\includegraphics[width=\textwidth]{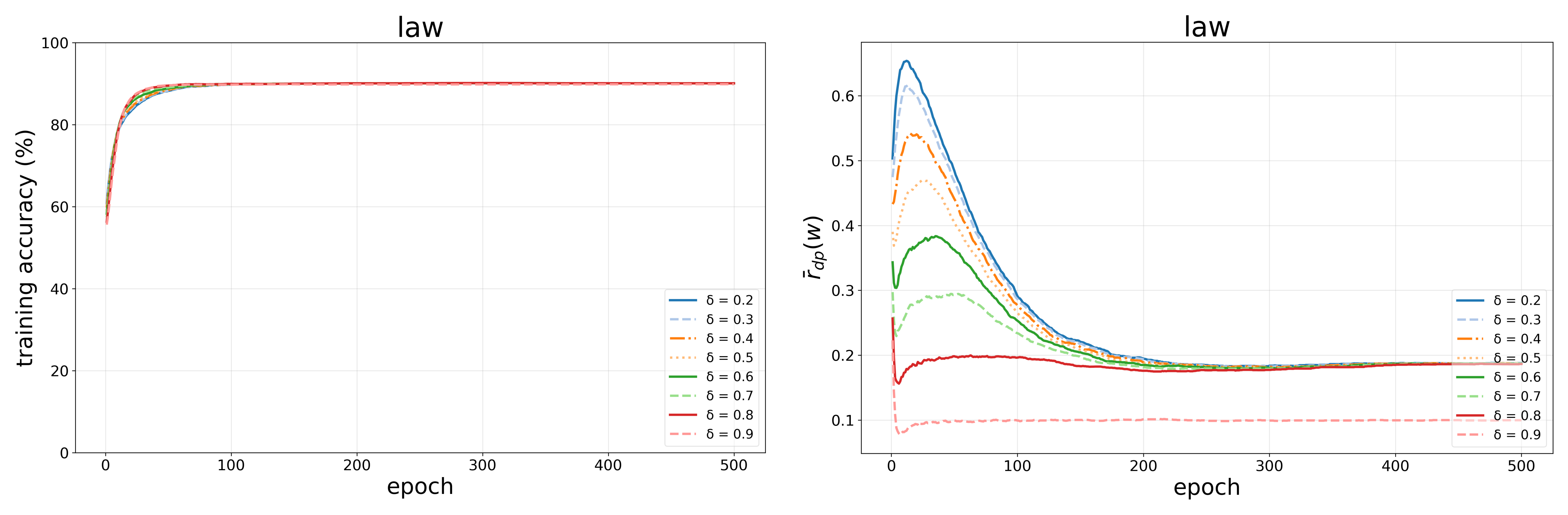}
\end{subfigure}

\caption{Comparison between the constraint-only (two right columns) and regularization-only (two left columns) approaches for enforcing fairness using the smoothed-step surrogate on Dutch and Law datasets. The constraint-only-based approach consistently meets the targeted threshold ($\delta$) with minimal impact on accuracy, while the regularization-only-based method exhibits unpredictable outcomes and greater accuracy loss, underscoring an advantage of using explicit constraints.}
\label{fig:reg_const}
\end{figure}

\begin{table*}[ht]
  \centering
  \caption{Accuracies (overall, male, and female) and constraint violations when disparate impact (DI) and/or equal impact (EI) constraints are enforced with $\delta = 0.8$ for the Dutch, Law, and ACSIncome datasets. 
  Enforcing both constraints reduces overall accuracy in some cases, but the constraint-only method consistently satisfies both unfairness constraints, as indicated by non-positive constraint model violations ($c_{di}$ and $c_{ei}$) and non-positive constraint violation measures ($\bar{c}_{di}$ and $\bar{c}_{ei}$).}
  \label{tab:multi_constraints_0.8}
    {\small
    \begin{tabular}{llccccccc}
      \toprule
      \textbf{Dataset} & \textbf{Scenario} & \textbf{Acc} & \textbf{Male Acc} & \textbf{Female Acc} & \textbf{$c_{di}$} & \textbf{$c_{ei}$} & \textbf{$\bar{c}_{di}$} & \textbf{$\bar{c}_{ei}$} \\
      \midrule
      \multirow{3}{*}{Dutch} 
      & Only DI       & 78.004   & 77.4006  & 78.6038  &  0.0     & -0.1224  & -0.0023   & -0.1191 \\
      & Only EI       & 82.2555  & 78.6538  & 85.8363  &  0.1729  &  0.0     &  0.1711   & -0.0023 \\
      & Both DI-EI    & 78.004   & 77.4006  & 78.6038  &  0.0     & -0.1224  & -0.0023   & -0.1191 \\
      \midrule
      \multirow{3}{*}{Law} 
      & Only DI       & 90.095   & 92.3973  & 77.9039  & -0.0124  & -0.1075  & -0.0127   & -0.1082 \\
      & Only EI       & 90.0829  & 92.3973  & 77.8282  & -0.012   & -0.1073  & -0.012    & -0.1072 \\
      & Both DI-EI    & 90.089   & 92.3973  & 77.8661  & -0.0124  & -0.1076  & -0.0131   & -0.1082 \\
      \midrule
      \multirow{3}{*}{ACSIncome} 
      & Only DI       & 78.5975  & 78.2293  & 79.0123  &  0.0     & -0.1139  &  0.0001   & -0.1139 \\
      & Only EI       & 78.8775  & 78.3284  & 79.496   &  0.0549  & -0.0472  &  0.0548   & -0.0482 \\
      & Both DI-EI    & 78.5475  & 78.2198  & 78.9166  &  0.0     & -0.1135  & -0.0002   & -0.1128 \\
      \bottomrule
    \end{tabular}%
  }
\end{table*}

\FloatBarrier

{\small
\begin{table}[ht]
    \centering
\caption{Compute times for full-batch training with fairness constraints across different datasets. "Avg time" is minutes for one 500-epoch run (averaged over 8 hyperparameter settings each: regularization-based approaches use 8 $\lambda$ values from 0.0001 to 1000, constraint-based approaches use 8 $\epsilon$ or $\delta$ values from 0.2 to 0.9); "Max peak" is the largest memory use seen in any run; "Total CPU" is the overall core-time spent on each approach.}
    \label{tab:comp_budget}
    \resizebox{\linewidth}{!}{
    \begin{tabular}{llllccc}
    \toprule
    \textbf{Dataset} & \textbf{Model Type} & \textbf{Approach} & \textbf{Hyperparams} & \textbf{Avg Time (min)} & \textbf{Max Peak (MB)} & \textbf{Total CPU (h)} \\
    \midrule
    \multirow{6}{*}{ACSIncome}
        & smoothed-step & regularization & 8 $\lambda$ values & 21.27 & 2441.2 & 7.54 \\
        & smoothed-step & constraint & 8 $\delta$ values & 23.82 & 2565.9 & 8.17 \\
        & sigmoid & regularization & 8 $\lambda$ values & 20.25 & 2230.4 & 7.33 \\
        & sigmoid & constraint & 8 $\delta$ values & 49.84 & 2244.8 & 17.20 \\
        & covariance & regularization & 8 $\lambda$ values & 30.32 & 2315.8 & 12.91 \\
        & covariance & constraint & 8 $\epsilon$ values & 52.13 & 2226.0 & 18.50 \\
    \midrule
    \multirow{6}{*}{Dutch}
        & smoothed-step & regularization & 8 $\lambda$ values & 32.49 & 555.6 & 5.21 \\
        & smoothed-step & constraint & 8 $\delta$ values & 32.58 & 555.7 & 5.26 \\
        & sigmoid & regularization & 8 $\lambda$ values & 33.16 & 562.5 & 5.24 \\
        & sigmoid & constraint & 8 $\delta$ values & 32.77 & 561.2 & 5.31 \\
        & covariance & regularization & 8 $\lambda$ values & 33.20 & 589.3 & 5.30 \\
        & covariance & constraint & 8 $\epsilon$ values & 23.92 & 557.3 & 7.75 \\
    \midrule
    \multirow{6}{*}{Law}
        & smoothed-step & regularization & 8 $\lambda$ values & 6.84 & 270.1 & 1.51 \\
        & smoothed-step & constraint & 8 $\delta$ values & 7.21 & 366.9 & 1.54 \\
        & sigmoid & regularization & 8 $\lambda$ values & 6.75 & 289.6 & 1.50 \\
        & sigmoid & constraint & 8 $\delta$ values & 6.76 & 266.4 & 1.50 \\
        & covariance & regularization & 8 $\lambda$ values & 6.74 & 281.3 & 0.90 \\
        & covariance & constraint & 8 $\epsilon$ values & 6.77 & 282.8 & 0.90 \\
    \bottomrule
    \end{tabular}}
\end{table}}

\begin{table*}[ht]
  \centering
  \caption{Accuracies (overall, male, and female) and constraint violations when disparate impact (DI) and/or equal impact (EI) constraints are enforced with $\delta = 0.9$ for the Dutch, Law, and ACSIncome datasets. 
  Enforcing both constraints reduces overall accuracy in some cases, but the constraint-only method consistently satisfies both unfairness constraints, as indicated by non-positive constraint model violations ($c_{di}$ and $c_{ei}$) and non-positive constraint violation measures ($\bar{c}_{di}$ and $\bar{c}_{ei}$).}
  \label{tab:multi_constraints_0.9}
    {\small
    \begin{tabular}{llccccccc}
      \toprule
      \textbf{Dataset} & \textbf{Scenario} & \textbf{Acc} & \textbf{Male Acc} & \textbf{Female Acc} & \textbf{$c_{di}$} & \textbf{$c_{ei}$} & \textbf{$\bar{c}_{di}$} & \textbf{$\bar{c}_{ei}$} \\
      \midrule
      \multirow{3}{*}{Dutch} 
      & Only DI       & 75.7034   & 77.2097  & 74.2058  &  0.0     & 0.0324   & -0.0092   & 0.0445 \\
      & Only EI       & 82.0134   & 77.8778  & 86.1251  &  0.1563  & 0.0      &  0.1565   & -0.0009 \\
      & Both DI-EI    & 76.9116   & 74.4917  & 79.3176  &  0.0     & -0.0096  & -0.0043   & -0.0069 \\
      \midrule
      \multirow{3}{*}{Law} 
      & Only DI       & 89.8786   & 92.383   & 76.6175  &  0.0     & -0.0595  & -0.0002   & -0.06 \\
      & Only EI       & 90.089    & 92.3973  & 77.8661  &  0.087   & -0.0078  &  0.0872   & -0.0079 \\
      & Both DI-EI    & 89.8786   & 92.383   & 76.6175  &  0.0     & -0.0595  & -0.0002   & -0.06 \\
      \midrule
      \multirow{3}{*}{ACSIncome} 
      & Only DI       & 78.355    & 77.9839  & 78.7731  &  0.0     & -0.0523  &  0.0001   & -0.0525 \\
      & Only EI       & 78.62     & 78.2246  & 79.0654  &  0.0754  &  0.0     &  0.075    & -0.0001 \\
      & Both DI-EI    & 78.3575   & 77.9839  & 78.7784  &  0.0     & -0.0523  &  0.0      & -0.0524 \\
      \bottomrule
    \end{tabular}%
  }
\end{table*}

\begin{figure}[!htbp]
\centering

\begin{subfigure}[b]{0.48\textwidth}
    \centering
    \includegraphics[width=\textwidth]{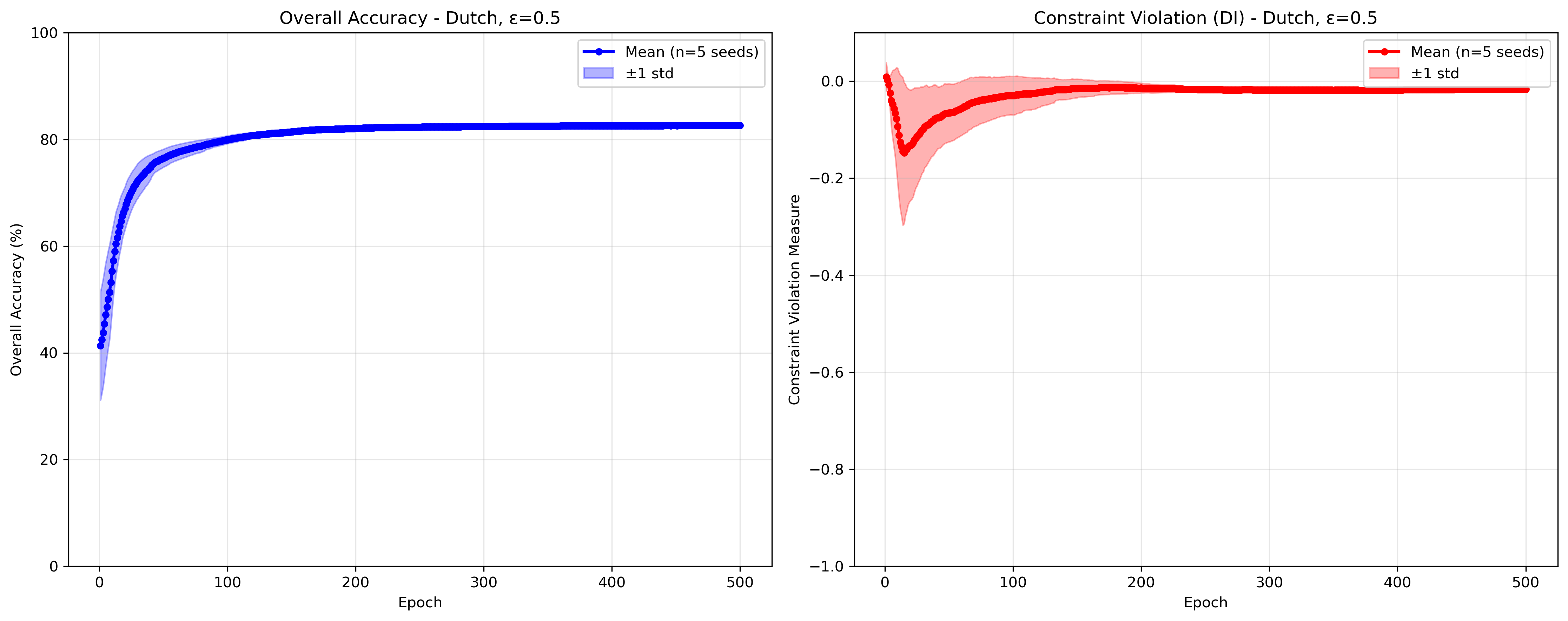}
    \caption{$\delta = 0.5$}
    \label{fig:dutch_eps_0_5}
\end{subfigure}
\hfill
\begin{subfigure}[b]{0.48\textwidth}
    \centering
    \includegraphics[width=\textwidth]{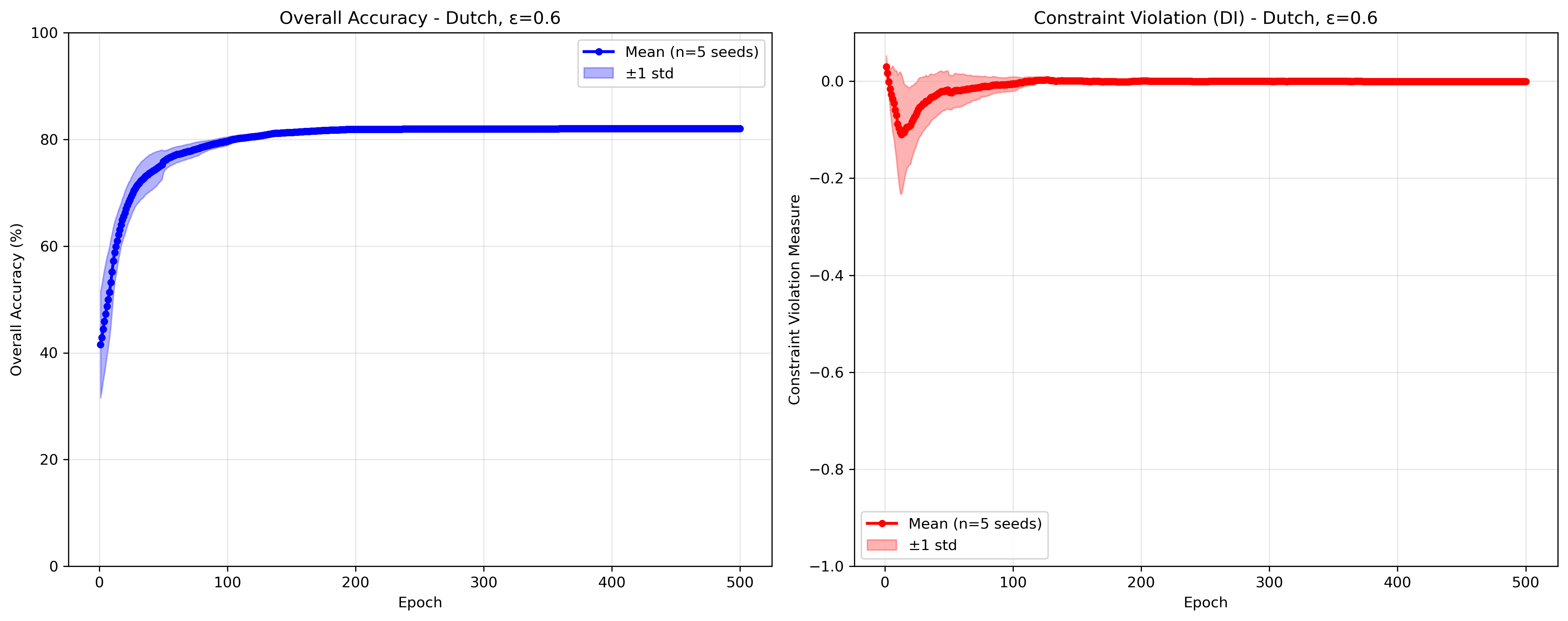}
    \caption{$\delta = 0.6$}
    \label{fig:dutch_eps_0_6}
\end{subfigure}

\vspace{0.2cm}

\begin{subfigure}[b]{0.48\textwidth}
    \centering
    \includegraphics[width=\textwidth]{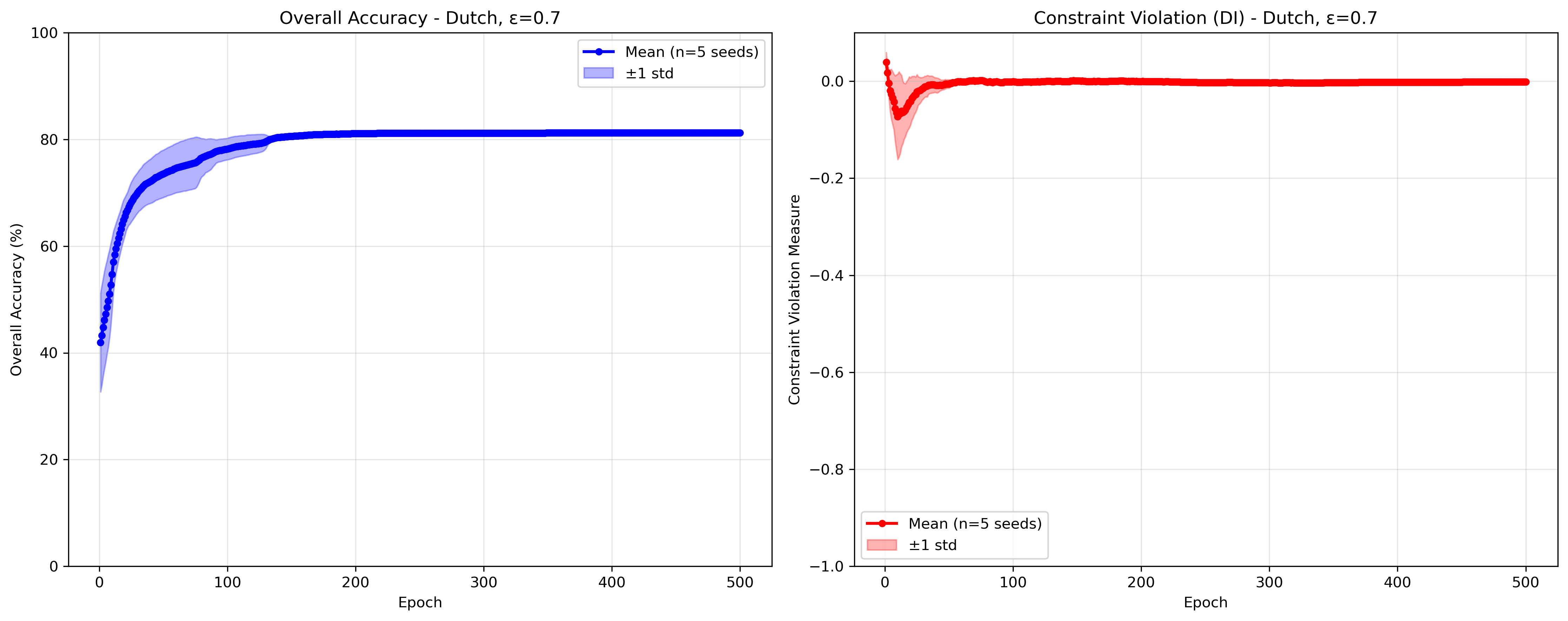}
    \caption{$\delta = 0.7$}
    \label{fig:dutch_eps_0_7}
\end{subfigure}
\hfill
\begin{subfigure}[b]{0.48\textwidth}
    \centering
    \includegraphics[width=\textwidth]{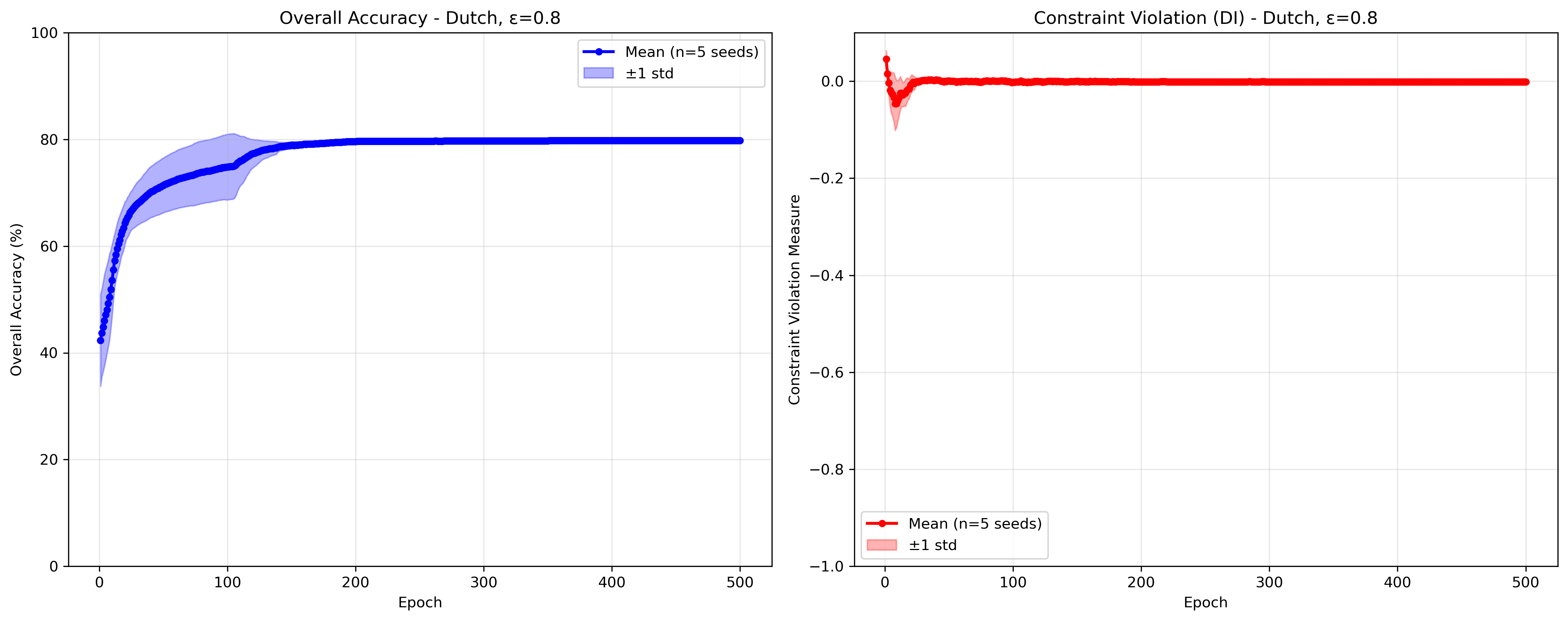}
    \caption{$\delta = 0.8$}
    \label{fig:dutch_eps_0_8}
\end{subfigure}

\vspace{0.2cm}

\begin{subfigure}[b]{0.48\textwidth}
    \centering
    \includegraphics[width=\textwidth]{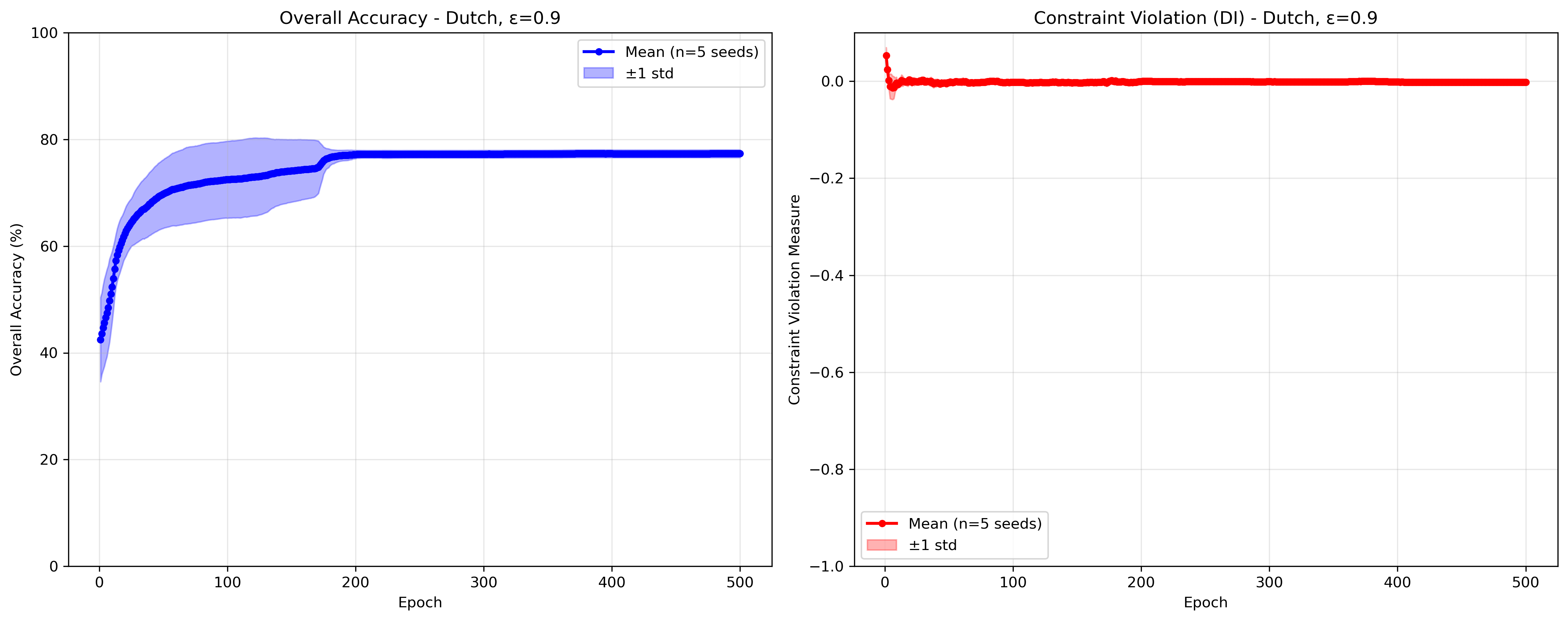}
    \caption{$\delta = 0.9$}
    \label{fig:dutch_eps_0_9}
\end{subfigure}

\caption{Variance analysis of training performance for Dutch dataset across different fairness levels. Left plots show overall accuracy over training epochs, right plots show constraint violation (DI) over training epochs. Solid lines represent the mean across 5 random seeds, while shaded regions indicate ±1 standard deviation. Each row corresponds to a different fairness parameter $\delta$, with lower values indicating stronger fairness constraint.}
\label{fig:dutch_training_plots}
\end{figure}

\begin{figure}[htbp]
\centering

\begin{subfigure}[b]{0.48\textwidth}
\centering
\includegraphics[width=\textwidth]{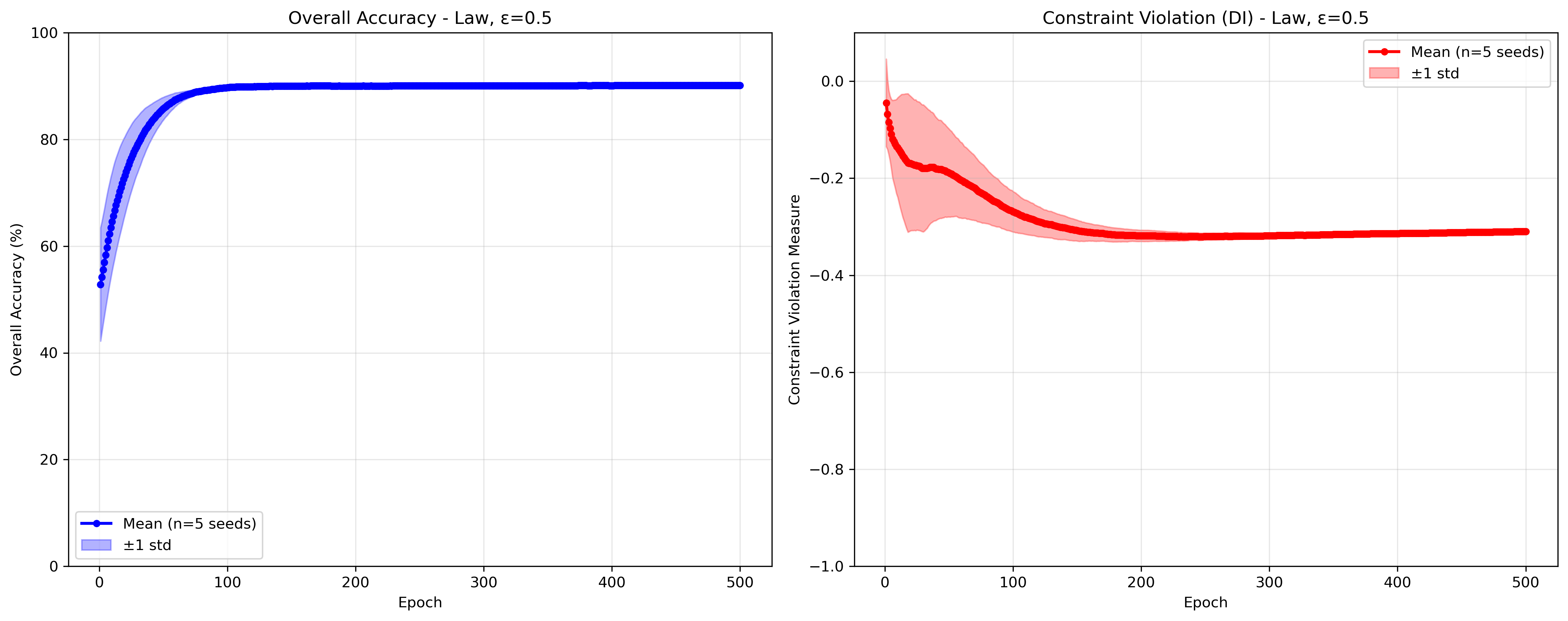}
\caption{$\delta = 0.5$}
\label{fig:law_eps_0_5}
\end{subfigure}
\hfill
\begin{subfigure}[b]{0.48\textwidth}
\centering
\includegraphics[width=\textwidth]{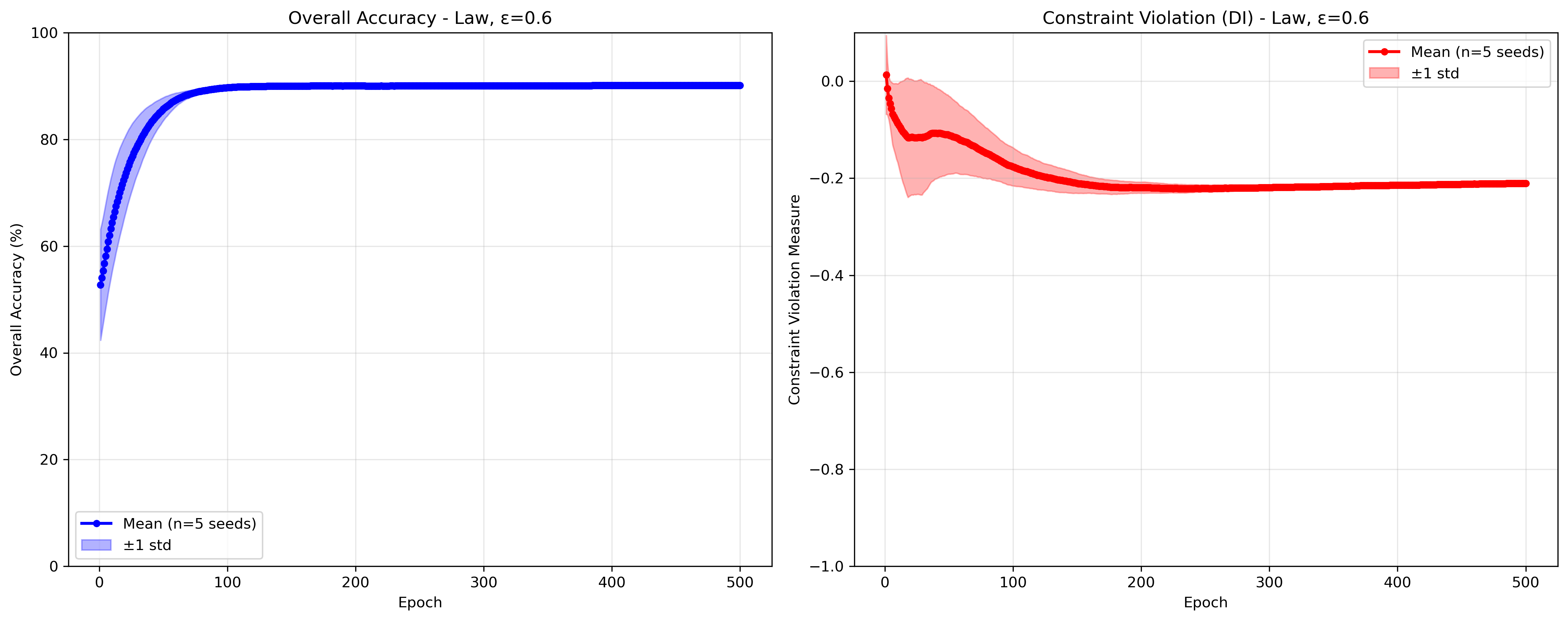}
\caption{$\delta = 0.6$}
\label{fig:law_eps_0_6}
\end{subfigure}

\vspace{0.25cm}

\begin{subfigure}[b]{0.48\textwidth}
\centering
\includegraphics[width=\textwidth]{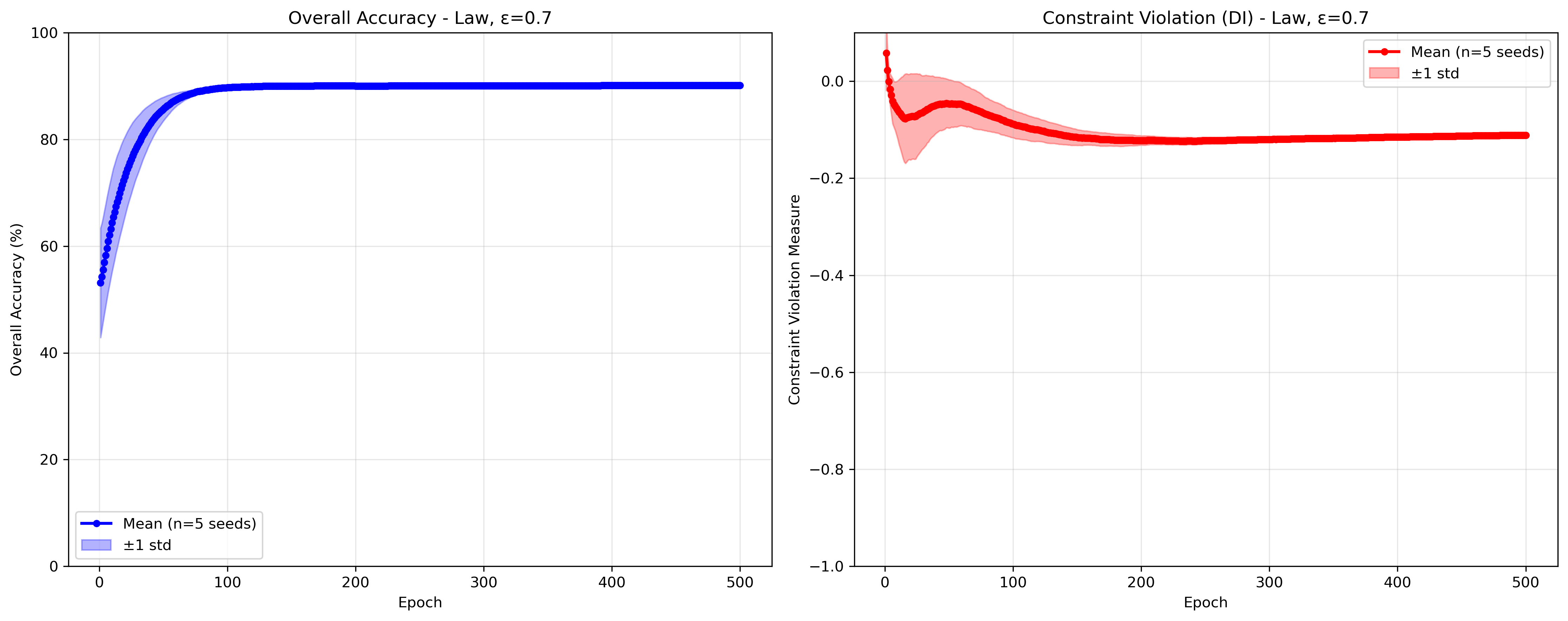}
\caption{$\delta = 0.7$}
\label{fig:law_eps_0_7}
\end{subfigure}
\hfill
\begin{subfigure}[b]{0.48\textwidth}
\centering
\includegraphics[width=\textwidth]{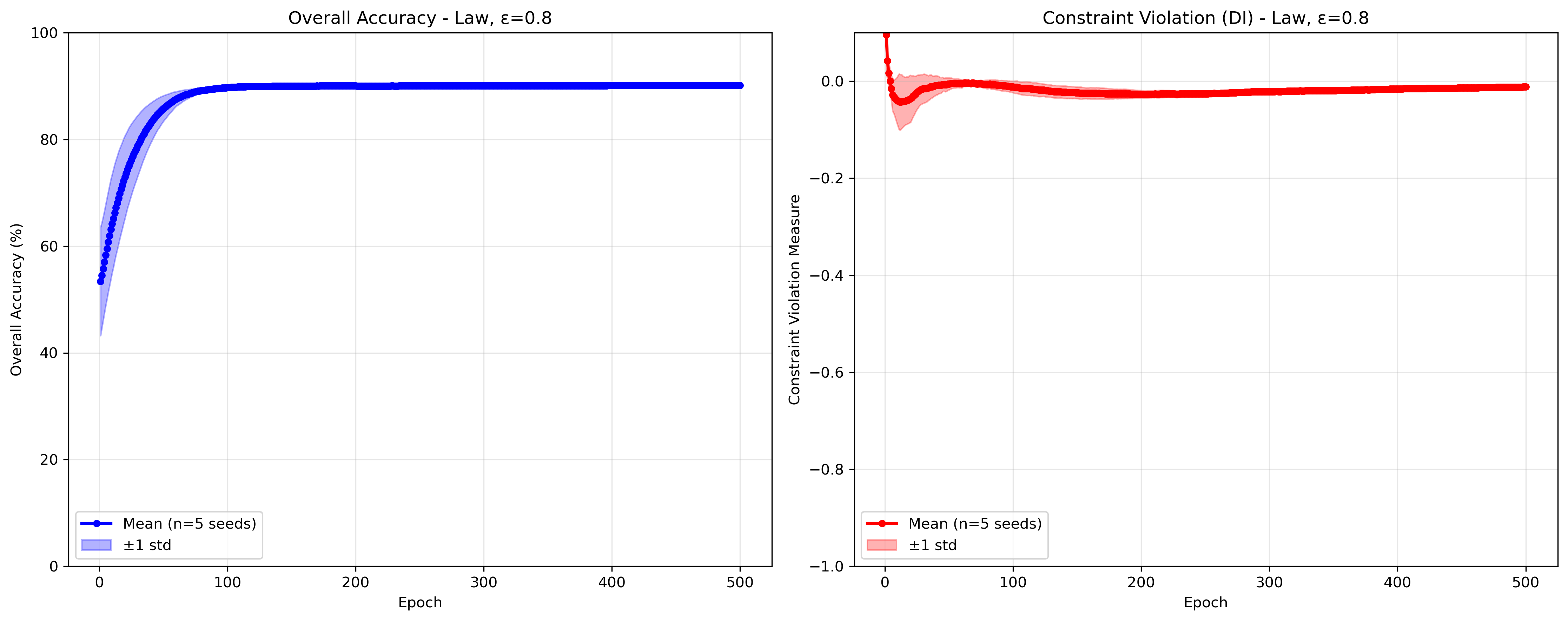}
\caption{$\delta = 0.8$}
\label{fig:law_eps_0_8}
\end{subfigure}

\vspace{0.25cm}

\begin{subfigure}[b]{0.48\textwidth}
\centering
\includegraphics[width=\textwidth]{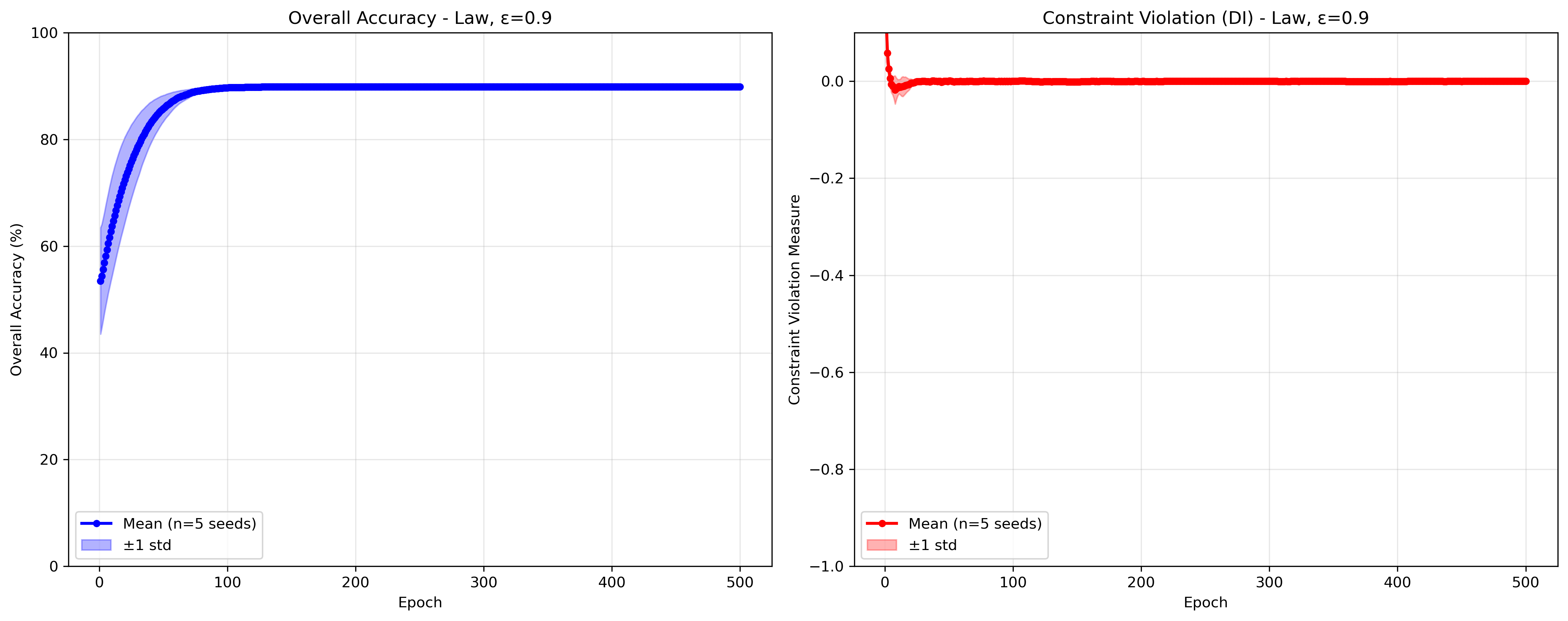}
\caption{$\delta = 0.9$}
\label{fig:law_eps_0_9}
\end{subfigure}

\caption{Variance analysis of training performance for the Law dataset across different fairness levels. 
Left plots show overall accuracy over training epochs, and right plots show constraint violation (DI) over training epochs. 
Solid lines represent the mean across 5 random seeds, while shaded regions indicate ±1 standard deviation. 
Each row corresponds to a different fairness parameter $\delta$, with lower values indicating stronger fairness constraints.}
\label{fig:law_training_plots}
\end{figure}

\begin{figure}[htbp]
\centering

\begin{subfigure}[b]{0.48\textwidth}
\centering
\includegraphics[width=\textwidth]{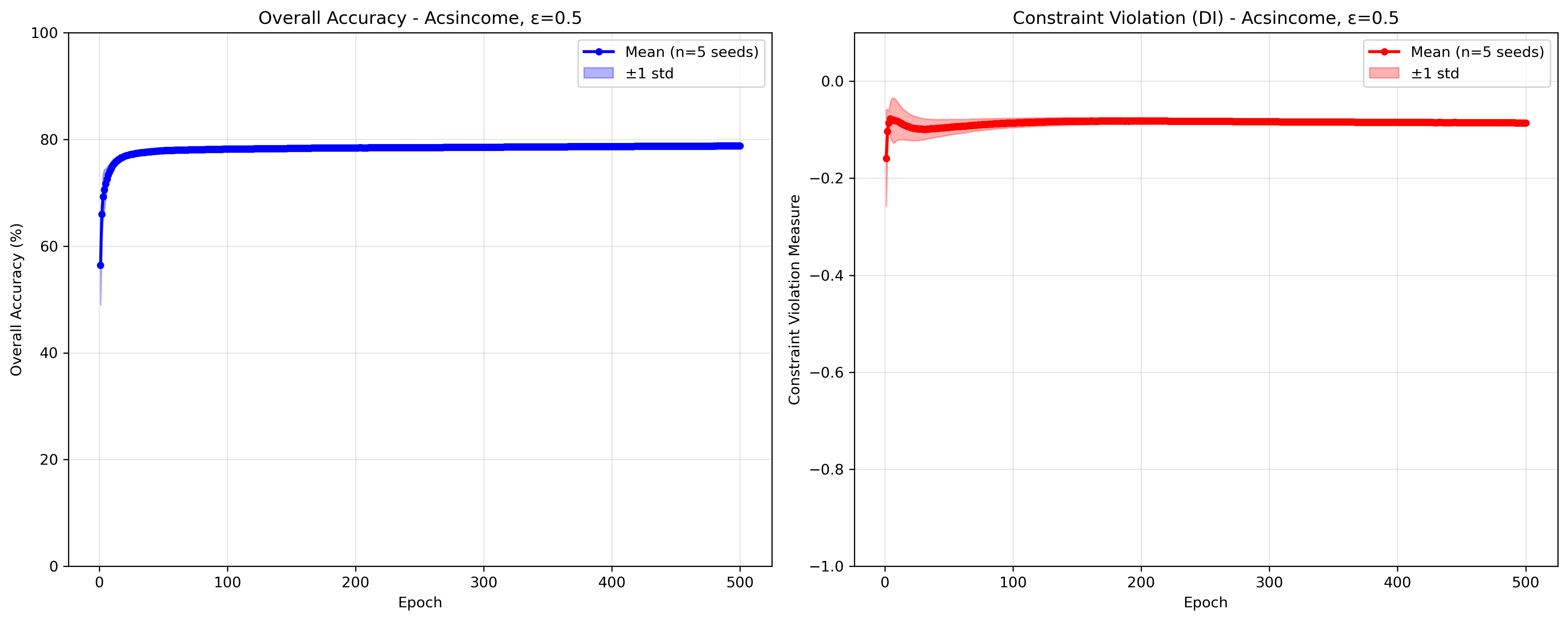}
\caption{$\delta = 0.5$}
\label{fig:acsincome_eps_0_5}
\end{subfigure}
\hfill
\begin{subfigure}[b]{0.48\textwidth}
\centering
\includegraphics[width=\textwidth]{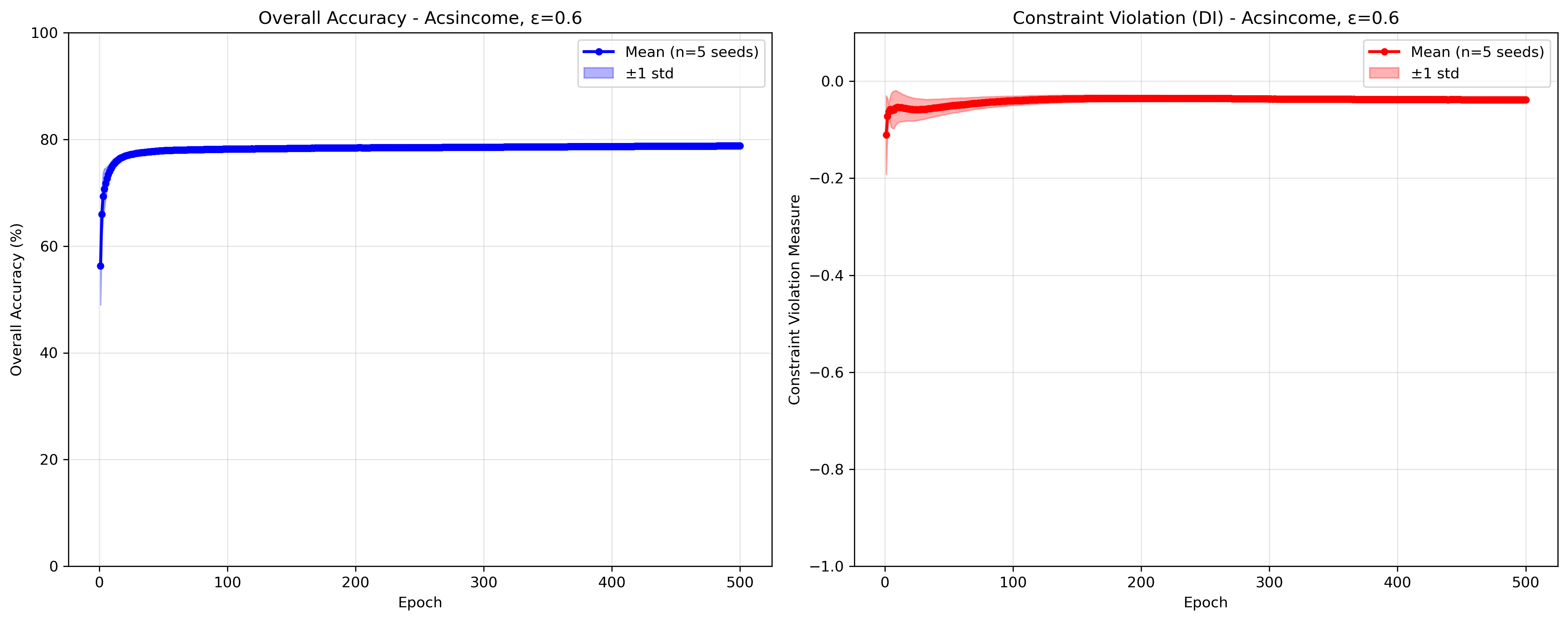}
\caption{$\delta = 0.6$}
\label{fig:acsincome_eps_0_6}
\end{subfigure}

\vspace{0.25cm}

\begin{subfigure}[b]{0.48\textwidth}
\centering
\includegraphics[width=\textwidth]{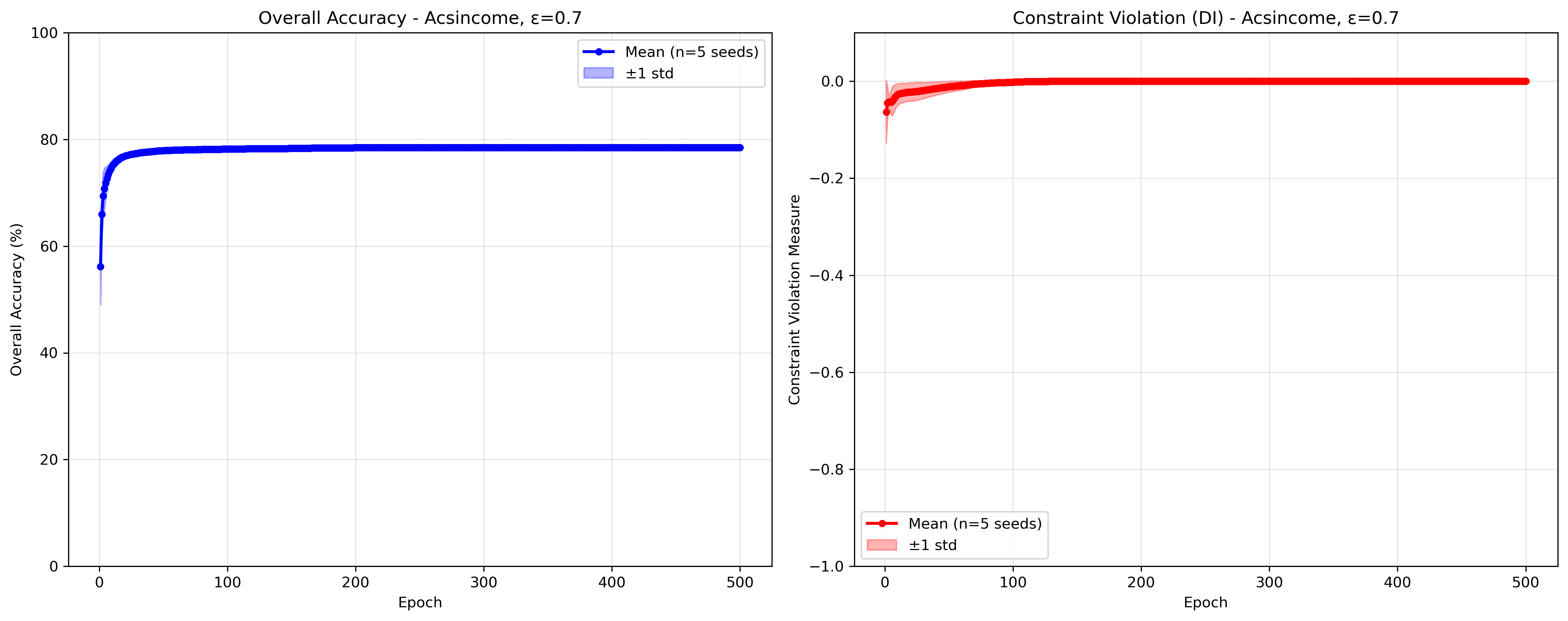}
\caption{$\delta = 0.7$}
\label{fig:acsincome_eps_0_7}
\end{subfigure}
\hfill
\begin{subfigure}[b]{0.48\textwidth}
\centering
\includegraphics[width=\textwidth]{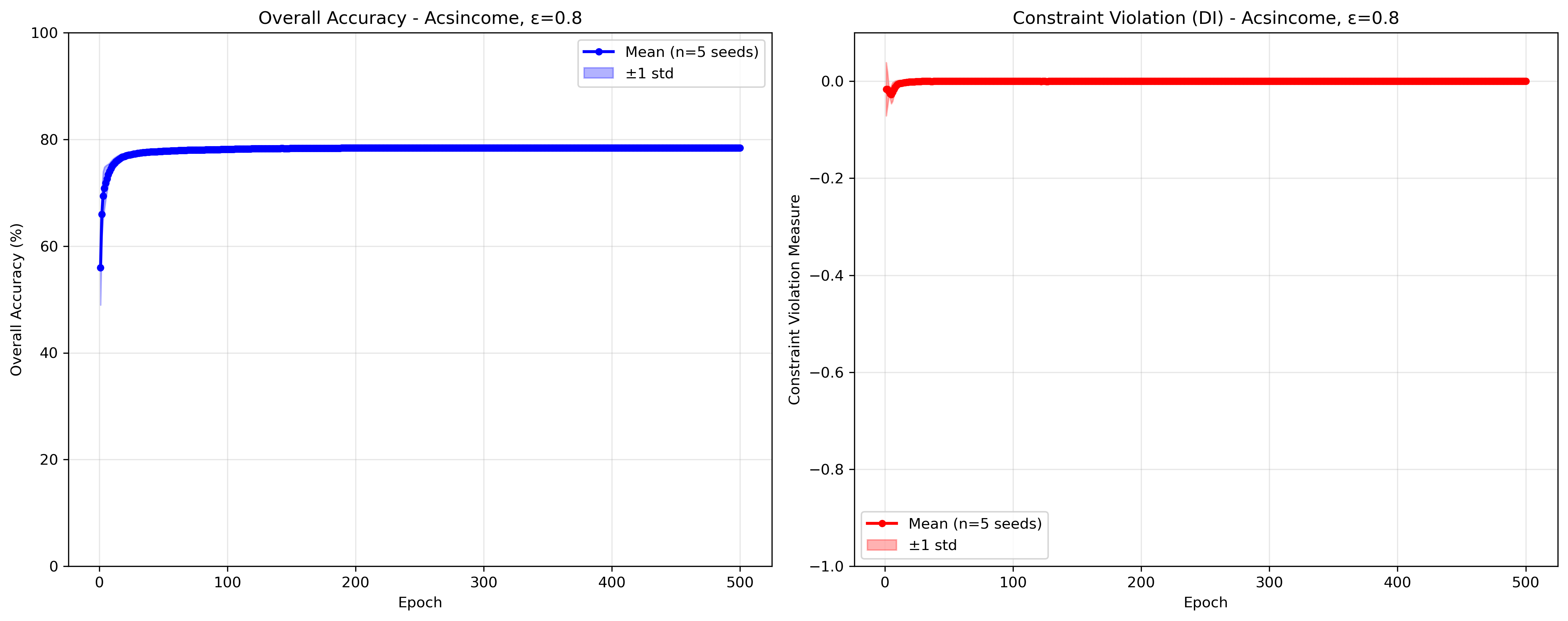}
\caption{$\delta = 0.8$}
\label{fig:acsincome_eps_0_8}
\end{subfigure}

\vspace{0.25cm}

\begin{subfigure}[b]{0.48\textwidth}
\centering
\includegraphics[width=\textwidth]{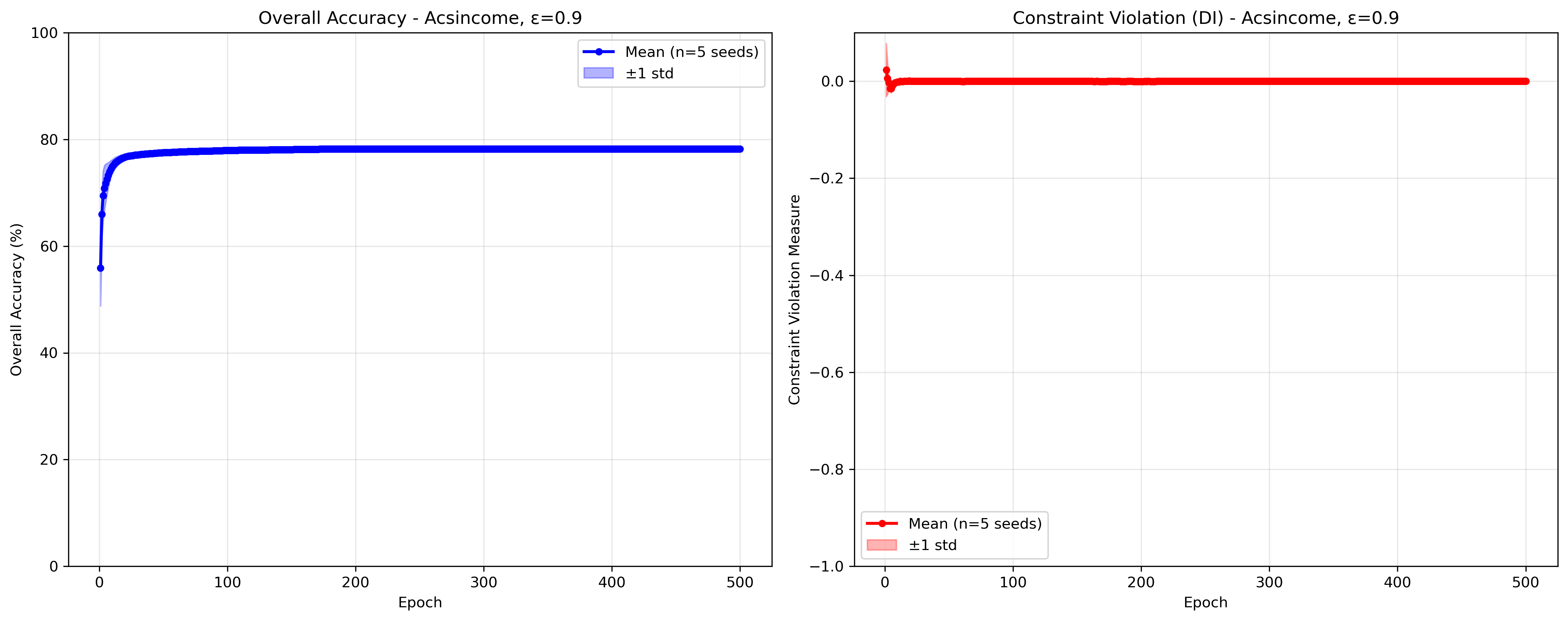}
\caption{$\delta = 0.9$}
\label{fig:acsincome_eps_0_9}
\end{subfigure}

\caption{Variance analysis of training performance for the ACSIncome dataset across different fairness levels. 
Left plots show overall accuracy over training epochs, and right plots show constraint violation (DI) over training epochs. 
Solid lines represent the mean across 5 random seeds, while shaded regions indicate ±1 standard deviation. 
Each row corresponds to a different fairness parameter $\delta$, with lower values indicating stronger fairness constraints.}
\label{fig:acsincome_training_plots}
\end{figure}